\documentclass[10pt,twocolumn,letterpaper]{article}

\usepackage[pagenumbers]{cvpr} 

\usepackage[dvipsnames]{xcolor}

\usepackage[utf8]{inputenc} 
\usepackage[T1]{fontenc}    
\usepackage{url}            
\usepackage{booktabs}       
\usepackage{amsfonts}       
\usepackage{nicefrac}       
\usepackage{microtype}      

\usepackage{graphicx}
\usepackage{amsmath}
\usepackage{amssymb}
\usepackage{booktabs}
\usepackage{wrapfig}

\usepackage{url}
\usepackage{epsfig}
\usepackage[accsupp]{axessibility}
\usepackage{helvet}   
\usepackage{courier}  
\usepackage{url}      
\usepackage{xcolor}
\usepackage{multirow}
\usepackage{subcaption}
\usepackage{algorithm}
\usepackage{algpseudocode}
\usepackage[export]{adjustbox}
\usepackage{color}
\usepackage{pifont}
\definecolor{mycolor}{RGB}{52, 128, 235}

\definecolor{cvprblue}{rgb}{0.21,0.49,0.74}
\usepackage[pagebackref,breaklinks,colorlinks,citecolor=cvprblue]{hyperref}

\def\paperID{731} 
\def\confName{CVPR}
\def\confYear{2024}

\title{Multi-Person 3D Pose Estimation from Multi-View Uncalibrated Depth Cameras}

\author{Yu-Jhe Li$^{1}$\qquad Yan Xu$^{2}$\qquad Rawal Khirodkar$^{3}$\qquad Jinhyung Park$^{1}$\qquad Kris Kitani$^{1,3}$\\
$^{1}$Carnegie Mellon University
\qquad
$^{2}$Waymo
\quad
$^{3}$Meta\\
{\tt\small \{\url{yujheli}, \url{kmkitani}\}\url{@andrew.cmu.edu}
}
}

\begin{document}
\maketitle

\begin{abstract}
We tackle the task of multi-view, multi-person 3D human pose estimation from a limited number of uncalibrated depth cameras. Recently, many approaches have been proposed for 3D human pose estimation from multi-view RGB cameras. However, these works (1) assume the number of RGB camera views is large enough for 3D reconstruction, (2) the cameras are calibrated, and (3) rely on ground truth 3D poses for training their regression model. In this work, we propose to leverage sparse, uncalibrated depth cameras providing RGBD video streams for 3D human pose estimation. We present a simple pipeline for Multi-View Depth Human Pose Estimation (MVD-HPE) for jointly predicting the camera poses and 3D human poses without training a deep 3D human pose regression model. This framework utilizes 3D Re-ID appearance features from RGBD images to formulate more accurate correspondences (for deriving camera positions) compared to using RGB-only features. We further propose (1) depth-guided camera-pose estimation by leveraging 3D rigid transformations as guidance and (2) depth-constrained 3D human pose estimation by utilizing depth-projected 3D points as an alternative objective for optimization. In order to evaluate our proposed pipeline, we collect three video sets of RGBD videos recorded from multiple sparse-view depth cameras, and ground truth 3D poses are manually annotated. Experiments show that our proposed method outperforms the current 3D human pose regression-free pipelines in terms of both camera pose estimation and 3D human pose estimation.

\end{abstract}

\section{Introduction}
\begin{figure*}[t!]
  \centering
  \includegraphics[width=0.85\linewidth]{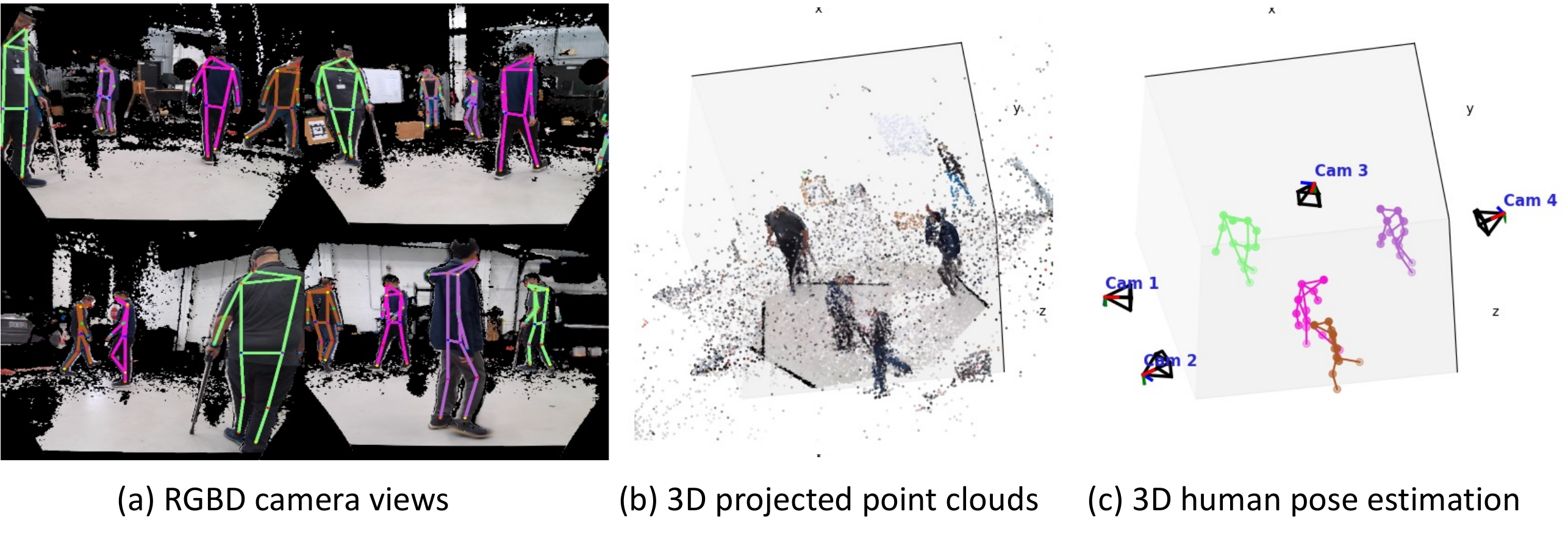}
  \vspace{-4mm}
  \caption{\textbf{3D human pose estimation from multi-view depth cameras}. Compared with multi-view RGB cameras, multi-view RGBD cameras provide additional depth information to reconstruct 3D point clouds more precisely.}
  \label{fig:teasor}
  \vspace{-4mm}
\end{figure*}

We approach the task of multi-view multi-person 3D human pose estimation from uncalibrated depth cameras. Recently, many approaches have been proposed for 3D human pose estimation from multi-view RGB cameras in either 3D regression-based~\cite{tu2020voxelpose,zhang2021direct,reddy2021tessetrack} or a 3D regression-free~\cite{belagiannis2014multiple,belagiannis20153d,ershadi2018multiple,dong2019fast} manners. Regression-based methods use a deep network trained on a specific 3D camera setup to estimate 3D poses. However, 3D regression-based models do not generalize well during inference to a different camera configuration or scene geometry. In contrast, regression-free methods detect 2D poses per camera view and perform triangulation to obtain the 3D poses. Both these approaches assume that (1) the camera poses are known and (2) the number of camera views is large enough for either regression using a deep network or reliable triangulation, which is often not applicable in the real world. Our key idea is to leverage the depth modality from a few uncalibrated cameras for explicit multi-view 3D reasoning, thereby reducing the number of necessary cameras, as shown in Figure~\ref{fig:teasor}.

Recently, approaches such as Wide-baseline~\cite{xu2021wide} and UncaliPose~\cite{Xu_2022_BMVC} have been proposed to resolve camera positions automatically by matching the appearance feature for human bodies of interest. UncaliPose~\cite{Xu_2022_BMVC} can perform 3D pose estimation jointly with camera pose estimation in a 3D regression-free manner. Though this approach can automatically estimate camera poses, later used for triangulation, the performance of camera pose estimation heavily relies on the 2D appearance feature for the cross-view association. As a result, when using RGB-only cameras, the 2D appearance features limit UncaliPose~\cite{Xu_2022_BMVC} from accurately resolving depth ambiguity and induce view-inconsistent errors. 3D reasoning abilities of a multi-view RGB camera system are severely restricted when the number of views is reduced; we believe rich 3D information from depth cameras can help address this issue.

To leverage several uncalibrated RGBD cameras for 3D human pose estimation, we propose a simple 3D regression-free pipeline for Multi-View Depth Human Pose Estimation (MVD-HPE) for jointly estimating the camera poses and 3D human poses. First, following previous 3D regression-free methods, we obtain 2D poses for all views using the off-the-shelf HRNet~\cite{sun2019deep}. Second, we extract human appearance features in 3D by utilizing the depth values from RGBD images and an off-the-shelf 3D Re-ID model~\cite{zheng2022parameter}. This allows us to formulate more accurate cross-view correspondences than those using RGB-only appearance features. Lastly, we propose 1) a depth-guided minimization objective for camera-pose estimation, which leads to more accurate camera calibration, and 2) a depth-constrained triangulation algorithm for accurate 3D human pose reconstruction. In order to evaluate our proposed pipeline, we collect a dataset containing several sets of RGBD videos recorded from multiple depth cameras. Furthermore, we manually annotated hundreds of frames as ground truth 3D poses for evaluation. The dataset along with the code will be released publicly upon acceptance.  To highlight the value of our uncalibrated system, it is designed for the scenario of smart cities where pedestrians are captured by either static or moving devices. Though the experimental datasets are currently recorded using depth cameras instead of Lidar+RGB, the pipeline that takes into colored point clouds from RGBD data is very applicable and adaptable to the outdoor scenarios with Lidar + RGB. To the best of our knowledge, we are among the first to design an uncalibrated system incorporating multi-view diverse sensors (RGB+Depth sensors). The contributions of this paper can be summarized as follows:

\begin{itemize}
\item We propose a simple regression-free approach called Multi-View Depth Human Pose Estimation (MVD-HPE) for 3D pose estimation. MVD-HPE works from a few uncalibrated cameras and leverages the depth modality for accurate cross-view association and 3D localization.

\item MVD-HPE employs our novel depth-guided minimization objective to estimate camera poses that are robust to errors in spatial correspondences.
\item We introduce a depth-constrained triangulation algorithm that leverages the constraint imposed by the 3D point clouds to reconstruct human poses accurately.

\item We evaluate MVD-HPE on our collected data showcasing its ability to accurately localize cameras in 3D and jointly estimate 3D poses for multiple humans. Our results show that MVD-HPE demonstrates superiority over existing regression-free models by a significant margin.
\end{itemize}

\section{Related Works}

\subsection{3D Human Pose Estimation}
Given images captured from multiple cameras, 3D human pose estimation in a multi-person setting can be performed using either single-view or multi-view images. Existing single-view methods can be divided into two categories: top-down methods, and bottom-up methods.
Top-down methods~\cite{rogez2017lcr,rogez2019lcr,zanfir2018monocular,moon2019camera,benzine2020pandanet} detect 2D human bounding boxes and perform 2D-to-3D pose lifting~\cite{chen20173d,martinez2017simple,li2019generating} or direct regression~\cite{li2015maximum,pavlakos2018ordinal,pavlakos2017coarse} to obtain 3D poses. Bottom-up methods~\cite{zanfir2018deep,nie2019single,fabbri2020compressed,mehta2018single,mehta2020xnect} estimate 3D joint locations for humans and then associate these joints with each human. Since these methods only use a single view, they do not need to address the multi-person cross-view association problem, but with a compromise of depth estimation ambiguity. On the other hand, multi-view methods can use information from multiple cameras for better localization in the 3D world. These methods also generally fall into two categories: 3D regression-free and 3D regression-based. 3D regression-free approaches are usually multi-stage~\cite{belagiannis2014multiple,belagiannis20153d,ershadi2018multiple,dong2019fast}. They first obtain the 2D poses~\cite{li2019rethinking,sun2019deep,cao2017realtime,cheng2020higherhrnet} before matching the 2D poses using appearance features~\cite{xu2021wide,xu2022joint} and geometry cues~\cite{dong2019fast}. Later they predict the 3D pose using multi-view geometry~\cite{hartley1997triangulation,burenius20133d}. 3D Regression-based approaches~\cite{tu2020voxelpose,zhang2021direct,reddy2021tessetrack} are usually single-stage and solve the problem using end-to-end regression. These methods first divide the scene into 3D voxels and localize each person from multi-view input. They then perform a fine-grained regression to obtain 3D joint locations. Recent state-of-the-art methods~\cite{reddy2021tessetrack,wu2021graph} utilize graph convolutional neural networks~\cite{kipf2016semi} and transformers~\cite{vaswani2017attention} for improved 3D reasoning. Compared to regression-free methods, regression-based methods require ground truth 3D poses for regressor training.
This greatly limits the application under in-the-wild settings where 3D labels are difficult to obtain. Considering this, we design our proposed approach, MVD-HPE, as a regression-free method that uses depth information from multiple camera views without the need for 3D labels.

\begin{figure*}[t!]
  \centering
  \includegraphics[width=\linewidth]{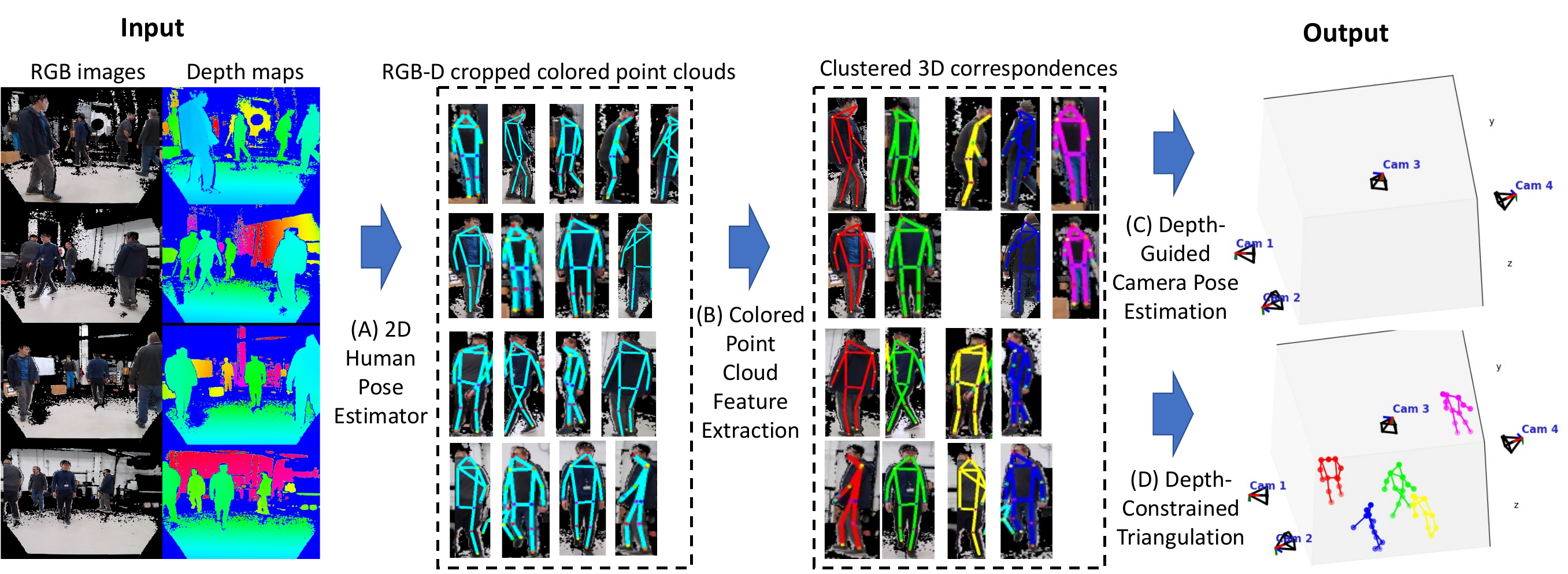}
  \caption{\textbf{Overview of the proposed pipeline for uncalibrated human pose estimation}. The pipeline contains four steps: a) 2D human pose estimation from the off-the-shelf 2D pose detector, b) colored point cloud feature extraction employing a 3D re-ID model, c) depth-guided camera pose estimation, and d) depth-constrained triangulation for 3D human pose estimation.}
  \label{fig:model}
\end{figure*}

\subsection{Camera Pose Estimation}
Camera pose estimation methods can be generally divided into two categories: geometric methods and deep camera pose regressors. Geometric methods consist of two stages, obtaining cross-view 2D correspondences and estimating the camera pose using 3D geometry. In the first stage, these methods detect key points (Harris~\cite{harris1988combined}, FAST~\cite{rosten2006machine}, etc.), describe them using hand-crafted features (SIFT~\cite{lowe2004distinctive}, BRIEF~\cite{calonder2010brief}, ORB~\cite{rublee2011orb}, etc.), and match them across images to obtain point correspondences ((BFM~\cite{jakubovic2018image}, FLANN \cite{muja2009flann}, etc.)). Recent deep learning methods ((SuperPoint~\cite{detone2018superpoint}, UR2KiD~\cite{yang2020ur2kid}, D2-net~\cite{dusmanu2019d2}, LIFT~\cite{yi2016lift}, Lf-net~\cite{ono2018lf}, Elf~\cite{benbihi2019elf})) detect and match key points simultaneously using neural networks. In the second stage, geometric methods estimate camera poses, typically using the N-point algorithm~\cite{hartley2003multiple, li2006five, nister2004efficient} to compute the essential matrix, which is then decomposed into a relative camera rotation and an up-to-scale translation~\cite{georgiev2014practical}. After that, bundle adjustment~\cite{triggs1999bundle} is often used to further refine the rotation and translation. With time, geometric methods have matured but struggle to match key point features across camera views when the distance between cameras is considerably large. Deep camera pose regressors were first applied to absolute camera pose estimation in \cite{kendall2015posenet,xu2020estimate}, which used a convolutional neural network to directly regress the camera pose~\cite{gu2018recent}. Lately, convolutional networks have also been applied to relative camera pose estimation in \cite{melekhov2017relative, laskar2017camera, balntas2018relocnet, ding2019camnet}. Despite the convenience of end-to-end regression, deep regressors~\cite{shavit2019introduction} still perform worse than geometric methods and require training data collection and labeling for each new scene. Therefore, they are not well suited for static camera tasks, which is the focus of this work. Recently, Wide-Baseline~\cite{xu2021wide} and UncaliPose~\cite{Xu_2022_BMVC} jointly solve the camera pose along with other perception tasks like tracking or human pose estimation by matching humans across views to obtain correspondences. MVD-HPE also follows a similar strategy to jointly reason about the camera positions and human poses but first introduces a scheme for using RGBD data for this task. Recently, OG-Net~\cite{zheng2022parameter} used 3D features for re-ID using meshes constructed using synthetic colored point clouds. Similarly, in the context of this work, since we have the depth information that can be transformed into colored point clouds, we leverage ~\cite{zheng2022parameter} as our re-ID feature extractor in MVD-HPE for the generalization.

\section{Method}
\subsection{Overview}
Given multi-view RGBD images, we aim to estimate 3D human poses. Specifically, we have $V$ views which give us RGB images $X=\{x_v\}_{v=1}^{V}$, where $x_v \in \mathbb{R}^{H \times W \times 3}$ and depth images $Z=\{z_v\}_{v=1}^{V}$, where $z_v \in \mathbb{R}^{H \times W \times 1}$ for each time stamp. Let $K$ be the number of humans in the scene, we aim to reconstruct the 3D poses $\{P_{k}\}_{k=1}^{K}$, where $P_k \in \mathbb{R}^{J \times 3}$ and each human has $J$ body key points.

The overview of our proposed method is presented in Figure~\ref{fig:model}. Our approach consists of four steps which include a) 2D human pose estimation, b) colored point cloud feature extraction, c) depth-guided camera pose estimation, and d) depth-constrained triangulation. In step (a), we detect bounding boxes and 2D poses from each view by using YOLOv3~\cite{redmon2018yolov3} and HRNet~\cite{sun2019deep}). For step (b), we use the 3D Re-ID model~\cite{zheng2022parameter} to extract 3D visual features from the RGBD images per detected bounding box. These features are then used to cluster (Sec.~\ref{sec:cluster}) humans, where each cluster represents cross-view correspondences of human body key points. For step (c), we perform camera pose estimation using these cross-view correspondences, where we leverage geometry constraints along with depth information as regularization (Sec.~\ref{sec:camera}). For step (d), we recover 3D human poses by triangulation using the predicted camera poses; additionally, we use depth consistency between the estimated key points and point cloud depth to refine our 3D pose estimates (Sec.~\ref{sec:triangulation}).

\subsection{Cross-View Matching with 3D Feature}\label{sec:cluster}

In order to jointly perform camera pose estimation and 3D pose estimation using triangulation, we not only need initial 2D key points but also their cross-view correspondences.


\begin{figure}[t!]
\centering
\includegraphics[width=0.9\linewidth]{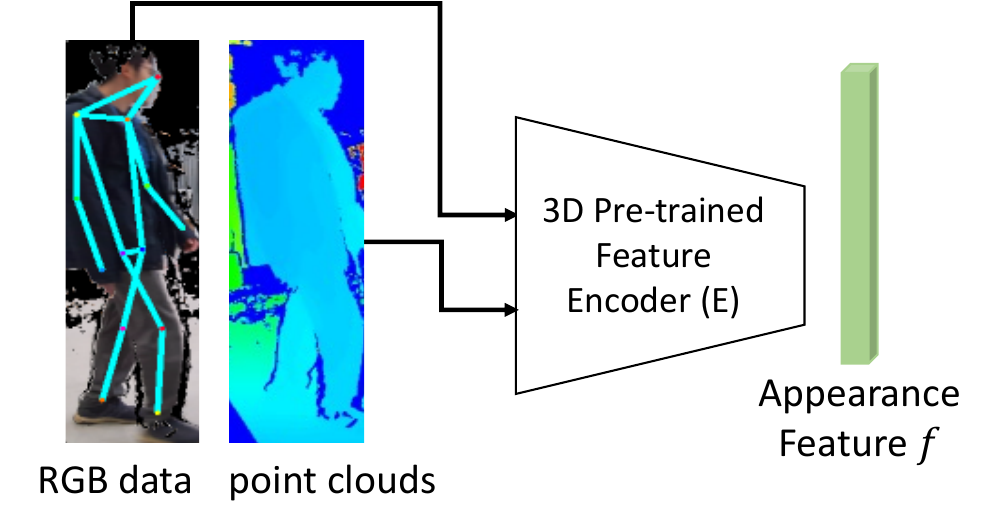}
\vspace{-5mm}
\caption{\label{fig:feature} Illustration of 3D appearance feature extraction taking into RGB image and depth image (transformed to point clouds).}
 \end{figure}

\begin{figure}[t!]
\centering
\includegraphics[width=\linewidth]{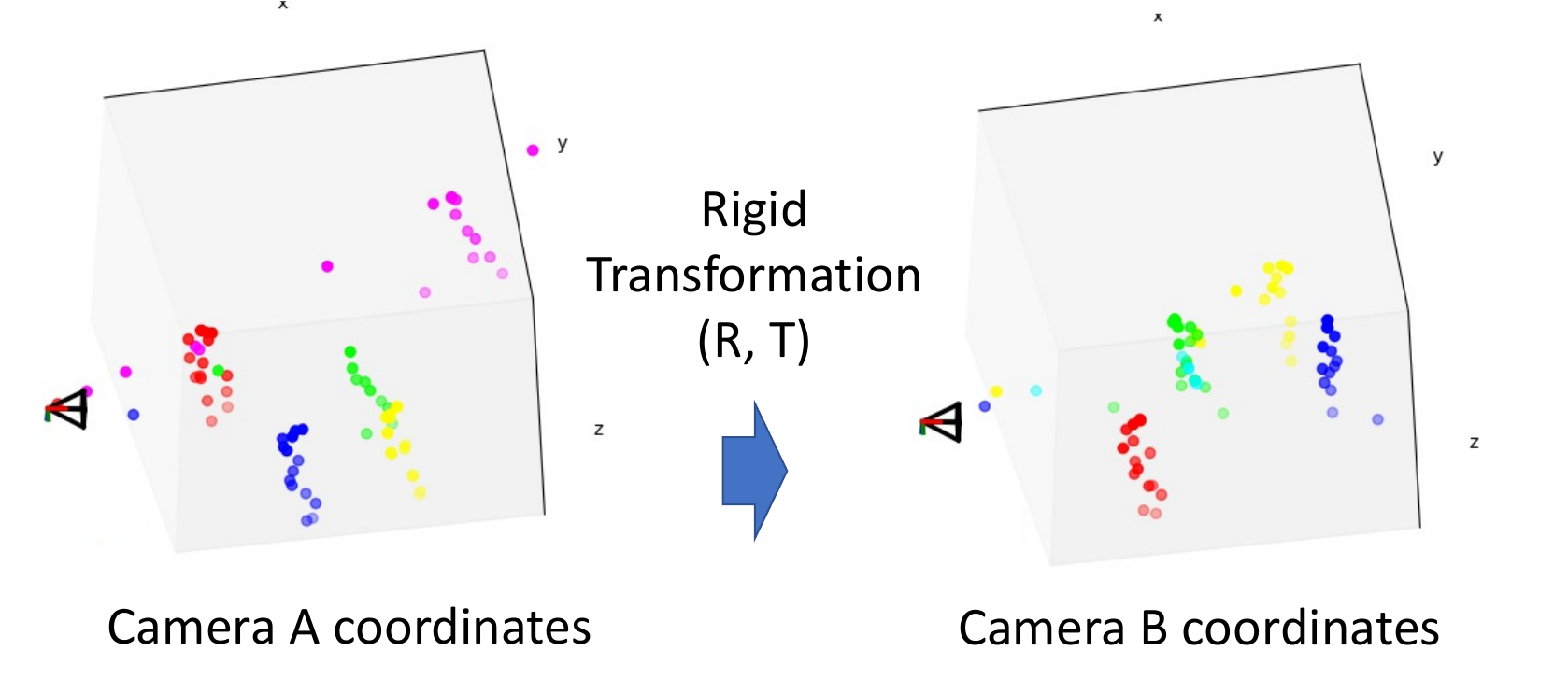}
\vspace{-7mm}
\caption{\label{fig:guide} Illustration of the rigid transformation (a rotation matrix $R$ and a up-to-scale translation $t$) can be resolved between two sets of 3D points. Not all of the 2D body key points can have depth measurements for 3D projected points.}
\end{figure}

Following ~\cite{dong2019fast,Xu_2022_BMVC,ershadi2018multiple,li2021visio}, we detect 2D body key points across all camera views usingYOLOv3~\cite{redmon2018yolov3} for bounding box detection and HRNet~\cite{sun2019deep} for top-down pose estimation. Note, as each depth camera may only observe $N_v$ humans, where $v \leq V$ is the camera index and $N_v \leq K$. As a result, we only have $N_v$ body key points $\{p_{v,n}\}_{v=1,n=1}^{V,N_v}$ where $p_{v,n} \in \mathbb{R}^{J \times 2}$.

To obtain cross-view body key point correspondences, we leverage a pretrained person re-identification model to extract features from each detected bounding box. While we acknowledge that re-ID models are invariant to illumination and viewpoint changes but most of them~\cite{dong2019fast,Xu_2022_BMVC,xu2021wide} are trained on massive image datasets and are not designed to exploit rich depth information from the RGBD images. For this reason, we leverage the recent 3D re-ID model~\cite{zheng2022parameter} trained on 3D colored point clouds obtained from synthetic meshes. The 3D re-ID model takes the RGBD information from the 2D bounding box as input and produces a one-dimensional feature vector as illustrated in Figure~\ref{fig:feature}. We denote these 3D appearance features as $\{f_{v,n}\}_{v=1,n=1}^{V,N_v}$ where $f_{v,n} \in \mathbb{R}^{512}$. Next, we cluster these features from all the views into $K$ groups $\{C^k | C^k \in \mathbb{R}^{512} \}_{k=1}^K$ as follows:
\begin{equation}
\begin{split}
\min_{c,w}~& \sum_{v=1}^V \sum_{n=1}^{N_v} \sum_{k=1}^{K} w_{v,n}^{k} \cdot (\|f_{v,n} - C^{k}\|_2^2)\\
s.t.~& \sum_{k=1}^{K} w_{v,n}^{k} = 1, w_{v,n}^{k} \in \{0,1\}\\
~& \text{(each feature is only assigned to one cluster)}\\
~& \sum_{v=1}^V \sum_{n=1}^{N_v} w_{v,n}^{k} \leq V \quad\\
~& \text{(the maximum number of features within one cluster)}\\
~&  \sum_{n=1}^{N_v} w_{v,n}^{k} \leq 1 \quad\\
~& \text{(same view constraint of features within one cluster)}\\
\end{split}
\label{eq:clustering}
\end{equation}
where $w_{v,n}^k$ is the assignment to the identity $k$.
We solve this optimization problem using the E-M algorithm~\cite{moon1996expectation}. For each iteration, the clusters $\{C^k_t | C_t^k \in \mathbb{R}^{512} \}_{k=1}^K$ are updated as:
\begin{equation}
\begin{split}
C^k_{t+1} = 
\begin{cases}
      \frac{\sum_{v=1}^V \sum_{n=1}^{N_v} w_{v,n}^{k} \cdot f_{v,n}}{\sum_{v=1}^V \sum_{n=1}^{N_v} w_{v,n}^{k}} & \text{if $\sum_{v=1}^V \sum_{n=1}^{N_v} w_{v,n}^{k}>0$}\\
      C^k_{t} & \text{otherwise}.
\end{cases} 
\end{split}
\label{eq:up_c}
\end{equation}
Finally, the body key points cross-view correspondences can be obtained from each cluster as $\{p_{v,k}\}_{v=1,k=1}^{V,K}$ where $p_{v,k} \in \{\mathbb{R}^{J \times 2},null\}$ and $p_{v,k}=null$ indicates no matched $k$ person in $v$ view. Specifically, the correspondences can be formulated as the tuple ($p_{i,k}$,$p_{j,k}$) given $i\neq j$, and both key points are not from null space.

\subsection{Depth-guided Camera Pose Estimation}\label{sec:camera}

In order to estimate the camera pose for each RGBD camera, we directly exploit the traditional eight-point algorithm~\cite{hartley1997defense} to produce the essential matrix $E_{AB}$ from camera $A$ to camera $B$. The essential matrix also provides the relative extrinsics: rotation $R_{AB}$ and up-to-scale (denoting $\alpha$) translation $t_{AB}$. The scale $\alpha$ can be estimated if we know the distance between the two cameras. However, since the previous work~\cite{Xu_2022_BMVC} for uncalibrated 3D pose estimation solely relies on the RGB feature correspondences for camera pose estimation, it suffers from depth inconsistencies. Toward this end, we propose to leverage the additional depth information to guide the camera pose estimation as an additional constraint during optimization.

Specifically, given the correspondences between two views: $\{(p_{i,k}, p_{j,k}) | p_{i,k}, p_{j,k}\notin \{null\} \}_{k=1}^{K}$, we estimate the essential matrix $E_{ij}$ by minimizing $\min_{E} \bold{A}\cdot \bold{v}_{ij}^E$, where $\bold{v}_{ij}^E \in \mathbb{R}^9$ denotes the flattened 3x3 matrix $E_{ij}$ and $\bold{A}$ indicates the coefficients for each correspondent equation (see eight-point algorithm~\cite{hartley1997defense}). This equation can be solved linearly using singular value decomposition (SVD) and takes the last column of the conjugate transpose matrix of the $V$ matrix in SVD. To leverage the depth information, we formulate the 3D point correspondences with depths as: $\{(P_{i,k}, P_{j,k}) | P_{i,k}, P_{j,k} \notin \{null\} \}_{k=1}^{K}$ where $P_{i,k} \in \mathbb{R}^{J \times 3}$ (more details in the supplemental). The 3D points can be null if there are no measured depths for some joints. If the derived extrinsic from the essential matrix is correct, the rotation matrix $R_{ij}$ and the translation $t_{ij}$ need to be matched with the rigid transformation $R'_{ij}$, $t'_{ij}$ between $P_{i,k}$ and $P_{j,k}$ as shown in Figure~\ref{fig:guide}. While the translation is up-to-scale, we can still apply the angle constraint between rotation matrices and formulate the complete objective to compute the essential matrix as follows: 
\begin{equation}
\begin{split}
 \min_{E,w} ~ & \bold{A}\cdot \bold{v}_{ij}^E + \cos^{-1} ( tr(R_{ij} \cdot R_{ij}^{'T} )/2-0.5)  - \sum_{s=1}^{J*K} w_s \\
s.t. ~ & \|\bold{v}_{ij}^E\|=1 \quad \text{(norm to be unit)}\\
& w_s \in \{0,1\} \quad\\
~& \text{(selected correspondence pairs among $J*K$ pairs)}\\
\textrm{where} ~ &  R_{ij} = \text{Decompose R from E} ~ (E)\\
~ &  R_{ij}^{'T} = VU^T \quad\\
~& \text{by} \quad [U,S,T] = SVD(\\
~& (P_{i,k}-centroid_{P_{i,k}})(P_{j,k}-centroid_{P_{j,k}}))\\
~ & (\text{multiply 3rd column of V by -1 if det(R) < 0}\\ 
~& \text{where the centroid indicates the mean})\\
\end{split}
\label{eq:camera_pose}
\end{equation}
We aim to utilize the maximum number of correct correspondences to solve the objective and compute the essential matrix. By solving the above objective, we obtain the camera extrinsics for all pairs of cameras: $M_{ij} = [R_{ij}|t_{ij}]$.

\begin{figure}
\centering
\vspace{-5mm}
\includegraphics[width=\linewidth]{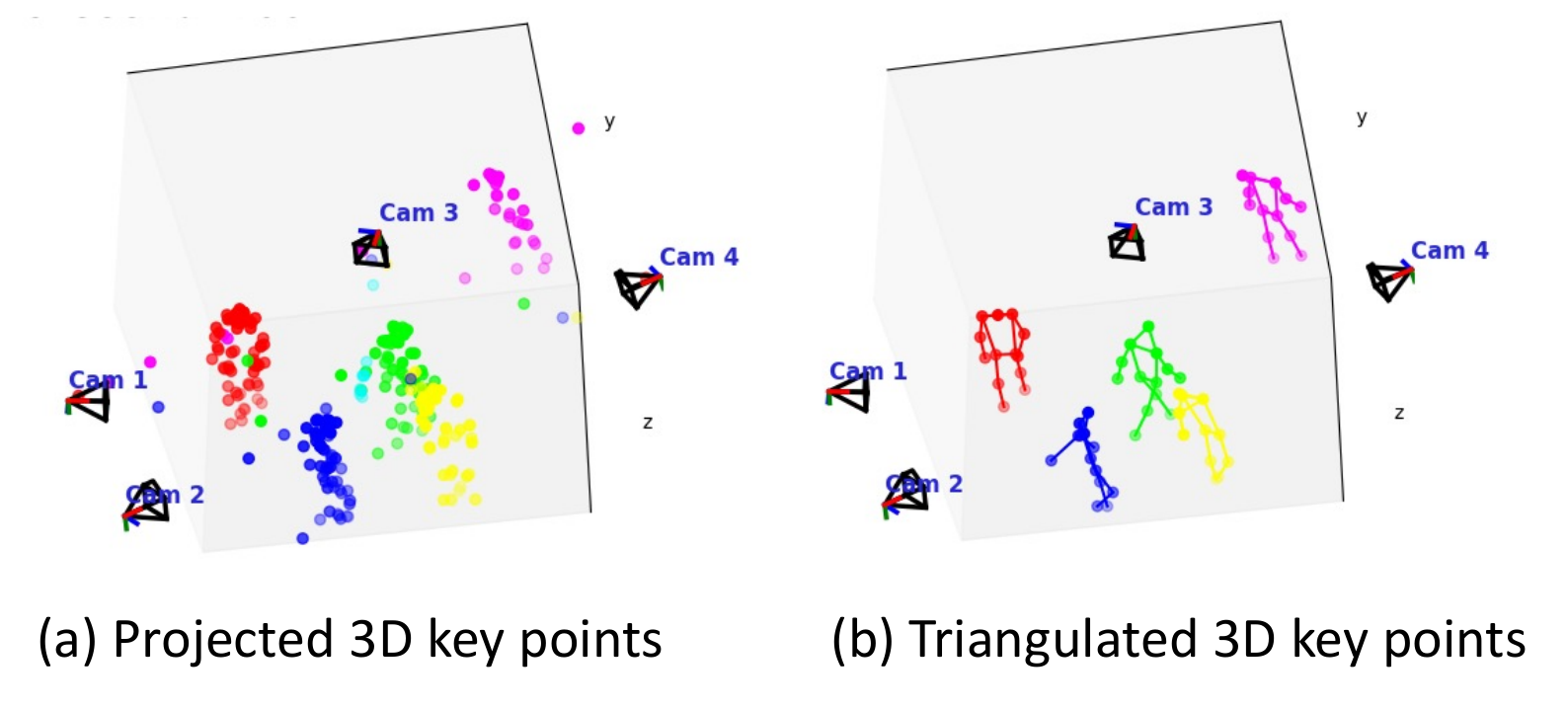}
\vspace{-6mm}
\caption{\label{fig:constraint} Illustration of (a) projected 3D key points from depths and (b) triangulation from the constrained objective. }
\end{figure}

\begin{figure*}[t!]
  \centering
  \includegraphics[width=0.8\linewidth]{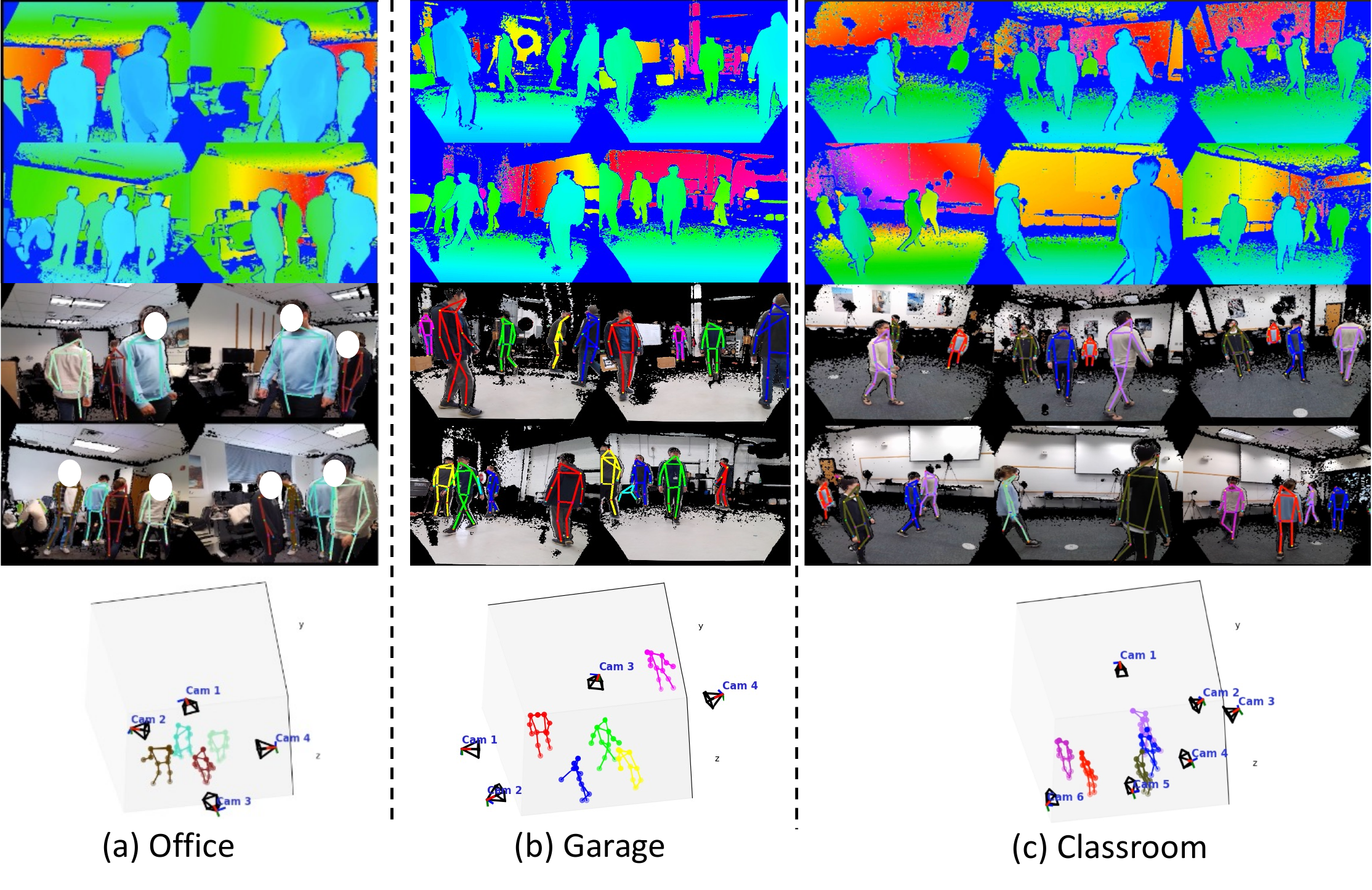}
  \vspace{-2mm}
  \caption{\textbf{Overview of the collected dataset}. We collected the data in three sites, including (a) office, (b) garage, and (c) classroom. The RGB images are downsampled, and the human faces are anonymized.}
  \label{fig:dataset}
\end{figure*}

\subsection{Depth-constrained 3D Human Pose Optimization}\label{sec:triangulation}
To obtain the 3D human poses, we perform triangulation on each pair of 2D point correspondences $\{(p_{i,k}, p_{j,k})\}_{k=1}^{K}$ with the computed camera matrices from the previous step. Additionally, to filter out the noisy 3D estimates, we impose constraints using the human body prior~\cite{burenius20133d} like constant bone length and left-right symmetry. However, these priors are not capable of fixing the initial triangulation errors due to incorrect cross-view correspondences.

To mitigate this issue, we leverage the depth information from each 2D correspondence and formulate the 3D point correspondences with depths as $\{(P_{i,k}, P'_{j,k})\}_{k=1}^{K}$, where $P'_{j,k}$ is the transformed 3D points from $P_{j,k}$ using camera extrinsics $M_{ji}$, as shown in Figure~\ref{fig:constraint}(a). The reconstructed 3D joints $P_k^{(i,j)}$ need to be close to both the 3D points $(P_{i,k}, P'_{j,k})$ if measured depth exists. Since the 3D points generated from depths can also induce additional errors, we take the average of those projected 3D depth points whose distance from the triangulated points is below a threshold. Our depth-constrained triangulation objective is as follows:
\begin{equation}
\begin{split}
 \min_{P_{k,d}^{(i,j)}, w_{k,d}} ~ & \sum_{d=1}^J (\bold{A}\cdot P_{k,d}^{(i,j)} \\
 ~& + w_{k,d} \cdot \|P_{k,d}^{(i,j)} - P_{k,d}^{depth}\|) - \sum_{d=1}^J  w_{k,d}\\
 s.t. ~ & w_{k,d} \in \{0,1\} \quad \text{(selected depth points}\\
 ~& \text{ and $w_{k,d} \equiv 0$ if $P_{k,d}^{depth}$ does not exist)}\\
\textrm{where} ~ & P_{k}^{depth} = \textrm{mean}((P_{i,k}, P'_{j,k}))\\
\end{split}
\label{eq:triangulation}
\end{equation}
$d$ indicates the index of the body joint. After obtaining all the candidates of the 3D joints using our revised objective for triangulation, we can either further apply the 3D pictorial structure prior to selecting the best point or take the average of the candidate's joints, as shown in Figure~\ref{fig:constraint}(b).

\section{Experiments}

\subsection{Dataset}

In order to evaluate our proposed method MVD-HPE for uncalibrated 3D human pose estimation, we collected data using multiple depth cameras (Microsoft Kinect) on three sites: 1) Office, 2) Garage, and 3) Classroom, as displayed in Figure~\ref{fig:dataset}. Each set of videos contains around 10k frames in total. We use the April tag~\cite{olson2011apriltag} and iterative closest point (ICP)~\cite{chetverikov2002trimmed} for ground-truth camera poses. \textbf{Office} videos were recorded in a cluttered environment with four people walking in the area with four cameras. \textbf{Garage} videos were recorded in an open garage with four people walking in the area and one fake person (doll) standing around along with four cameras. \textbf{Classroom} videos were recorded indoors in a conference room with five people walking around the area with six cameras. We manually annotate 200 frames with 3D ground truth poses from each site for evaluation. For data collection, we followed the standard guidelines from the IRB and obtained signed consent from all the participants for data release.

\subsection{Evaluation details}
For \textbf{camera pose estimation,} we report average angle error for rotation ($^{\circ}$) and translation error (mm) between ground truth and predicted camera poses across $T$ frames as the metric score. For \textbf{3D human pose estimation,} we follow the same evaluation protocols as previous works~\cite{Xu_2022_BMVC,dong2019fast} and report the percentage of correctly estimated parts (PCP) to measure the accuracy of the predicted 3D poses. For each video set, we pre-select four actors for evaluation. The "Classroom" set has five people, and we only choose four for evaluation. For a fair comparison, we reproduce all the prior art on our dataset.


\begin{table*}[t!]
\caption{\textbf{The results and comparisons of camera pose estimation.} We report the camera position (mm) and orientation angle ($^{\circ}$) errors for the predicted camera poses. For each dataset, we report the mean position and orientation errors of all cameras among the geometric methods (first big row), current methods (second big row), and our approaches (third big row).}
\label{tab:cam_pose_err}
\centering
  \begin{tabular}{l|c|c|c|c}
    \toprule
     & \multicolumn{3}{c|}{Camera pose error ($mm, {}^{\circ}$) $\downarrow$} \\
    \cmidrule(r){2-4}
    \hfil\multirow{-2}{*}{Method} & \multicolumn{1}{c}{\textit{Office}} & \multicolumn{1}{c}{\textit{Garage}} & \multicolumn{1}{c|}{\textit{Classroom}} & \multirow{-2}{*}{\parbox{2.0cm}{\centering Mean $\downarrow$}}\\
    \midrule
    SIFT~\cite{lowe2004distinctive} + BFM~\cite{jakubovic2018image} & $4346mm, 53.77^{\circ}$ & $6024mm, 37.18^{\circ}$ & $7037mm, 45.34^{\circ}$ & $5802mm, 45.43^{\circ}$\\
    SuperPoint~\cite{detone2018superpoint} + BFM~\cite{jakubovic2018image} & $3021mm, 44.93^{\circ}$ & $4761mm, 53.42^{\circ}$ & $4491mm, 52.38^{\circ}$ & $4091mm, 50.24^{\circ}$ \\
    \midrule
    Wide-Baseline~\cite{xu2021wide} & $586mm, 2.41^{\circ}$ & $540mm, 2.28^{\circ}$ & $511mm, 1.97^{\circ}$ & $546mm, 2.22^{\circ}$\\
    Wide-Baseline~\cite{xu2021wide} + ICP~\cite{chetverikov2002trimmed} & $421mm, 2.21^{\circ}$ & $376mm, 1.54^{\circ}$ & $439mm, 1.66^{\circ}$ & $412mm, 1.80^{\circ}$\\
    UncaliPose~\cite{Xu_2022_BMVC} & $397mm, 1.91^{\circ}$ & $422mm, 1.78^{\circ}$ & $451mm, 2.64^{\circ}$ & $423mm, 2.11^{\circ}$\\
    UncaliPose~\cite{Xu_2022_BMVC} + ICP~\cite{chetverikov2002trimmed} & $335mm, 1.23^{\circ}$ & $278mm, 0.89^{\circ}$ & $318mm, 1.97^{\circ}$ & $310mm, 1.36^{\circ}$ \\
    \midrule
    MVD-HPE w/o depth-guide & $365mm, 1.78^{\circ}$ & $390mm, 1.28^{\circ}$ & $430mm, 2.13^{\circ}$ & $395mm, 1.73^{\circ}$ \\
    MVD-HPE w/o depth-guide (ideal matches) & $285mm, 0.91^{\circ}$ & $325mm, 1.25^{\circ}$ & $382m, 1.79^{\circ}$ & $330mm, 1.31^{\circ}$ \\
    MVD-HPE w/o depth-guide + ICP~\cite{chetverikov2002trimmed}  & $294mm, 0.88^{\circ}$ & $282mm, 0.77^{\circ}$ & $321mm, 1.25^{\circ}$ & $299mm, 0.97^{\circ}$ \\
    
    MVD-HPE (Ours) & $172mm, 1.24^{\circ}$ & \textbf{$232mm, 1.45^{\circ}$} & \textbf{$180mm, 1.12^{\circ}$} & $194mm, 1.27^{\circ}$ \\
    MVD-HPE (ideal matches) & $142mm, 0.75^{\circ}$ & \textbf{$209mm, 1.07^{\circ}$} & \textbf{$165mm, 1.15^{\circ}$} & $172mm, 0.99^{\circ}$ \\
    MVD-HPE + ICP~\cite{chetverikov2002trimmed} (Ours) & $\textbf{135mm}, \textbf{0.79}^{\circ}$ & $\textbf{118mm}, \textbf{0.27}^{\circ}$ & $\textbf{156mm}, \textbf{0.55}^{\circ}$ & $\textbf{136mm}, \textbf{0.53}^{\circ}$\\

    \bottomrule
  \end{tabular}

\end{table*}
\begin{table*}[t!]
\small
\caption{\textbf{Comparisons on our collected datasets: Office, Garage, and Classroom.} The reported numbers are PCP values. The number in bold indicates the best results. }
\label{tab:pcp_sota}
\centering
  \begin{tabular}{l|c|cccc|c}
    \toprule
    \multicolumn{1}{c}{Office} & \multicolumn{1}{|c}{Camera Pose} & \multicolumn{1}{|c}{Actor 1} & \multicolumn{1}{c}{Actor 2} & \multicolumn{1}{c}{Actor 3} & \multicolumn{1}{c}{Actor 4} & \multicolumn{1}{|c}{Average $\uparrow$}\\
    \midrule
     Belagiannis \etal~\cite{belagiannis20153d}& \checkmark & 85.2 & 77.1 & 75.2 & 71.9 & 77.4\\
     Ershadi \etal~\cite{ershadi2018multiple}& \checkmark & 88.4 & 80.1 & 79.5 & 77.3 & 81.3\\
     Dong \etal~\cite{dong2019fast}& \checkmark & 94.1 & 95.5 & 93.7 & 94.0 & 94.3\\
     UncaliPose~\cite{Xu_2022_BMVC} & - & 96.5 & 93.1 & 94.2 & 93.3 & 94.3\\
    \midrule
    MVD-HPE (Ours) & - & 98.2 & 96.8 & 98.2 & 95.9 & 97.3\\
    MVD-HPE (Ours) & \checkmark  & \textbf{99.5} & \textbf{97.1} & \textbf{98.9} & \textbf{96.2} & \textbf{97.9}\\
    \midrule
    \midrule
    \multicolumn{1}{c}{Garage} & \multicolumn{1}{|c}{Camera Pose} & \multicolumn{1}{|c}{Actor 1} & \multicolumn{1}{c}{Actor 2} & \multicolumn{1}{c}{Actor 3} & \multicolumn{1}{c}{Actor 4} & \multicolumn{1}{|c}{Average $\uparrow$}\\
    \midrule
     Belagiannis \etal~\cite{belagiannis20153d}& \checkmark & 74.4 & 84.3 & 85.2 & 81.6 & 81.4 \\
     Ershadi \etal~\cite{ershadi2018multiple}& \checkmark & 84.1 & 89.2 & 89.5 & 87.3 & 87.5 \\
     Dong \etal~\cite{dong2019fast}& \checkmark & 92.4 & 93.4 & 94.6 & 94.9 & 93.8\\
     UncaliPose~\cite{Xu_2022_BMVC} & - & 94.7 & 95.1 & 94.8 & 95.2 & 95.0\\
    \midrule
    MVD-HPE (Ours) & - & 97.1 & 96.2 & 96.9 & 96.5 & 96.7\\
    MVD-HPE (Ours) & \checkmark  & \textbf{98.1} & \textbf{97.5} & \textbf{97.4} & \textbf{97.1} & \textbf{97.5}\\
     \midrule
    \midrule 
    \multicolumn{1}{c}{Classroom} & \multicolumn{1}{|c}{Camera Pose} & \multicolumn{1}{|c}{Actor 1} & \multicolumn{1}{c}{Actor 2} & \multicolumn{1}{c}{Actor 3} & \multicolumn{1}{c}{Actor 4} & \multicolumn{1}{|c}{Average $\uparrow$}\\
    \midrule
     Belagiannis \etal~\cite{belagiannis20153d}& \checkmark & 80.5 & 73.1 & 80.7 & 77.1 & 77.9\\
     Ershadi \etal~\cite{ershadi2018multiple}& \checkmark & 86.4 & 79.5 & 87.7 & 84.7 & 84.6\\
     Dong \etal~\cite{dong2019fast}& \checkmark & 93.5 & 92.7 & 94.2 & 93.1 & 93.4\\
     UncaliPose~\cite{Xu_2022_BMVC} & - & 94.8 & 91.0 & 94.9 & 92.8 & 93.4\\
    \midrule
    MVD-HPE (Ours) & - & 97.7 & 96.2 & 98.9 & 96.0 & 97.2\\
    MVD-HPE (Ours) & \checkmark & \textbf{98.4} & \textbf{96.9} & \textbf{99.4} & \textbf{97.3} & \textbf{98.0}\\
    \bottomrule
  \end{tabular}
\end{table*}
\begin{table*}[t!]
\small
\caption{Ablation studies on 3D human pose estimation for depth-constrained triangulation (second big row) and depth-guided pose estimation (third big row). The reported numbers are PCP values. }
\label{tab:pcp_abl}
\centering
  \begin{tabular}{l|c|cccc|c}
    \toprule
    \multicolumn{1}{c}{Classroom} & \multicolumn{1}{|c}{Camera Pose} & \multicolumn{1}{|c}{Actor 1} & \multicolumn{1}{c}{Actor 2} & \multicolumn{1}{c}{Actor 3} & \multicolumn{1}{c}{Actor 4} & \multicolumn{1}{|c}{Average $\uparrow$}\\
    \midrule
    
    MVD-HPE (Ours) & - & 97.7 & 96.2 & 98.9 & 96.0 & 97.2\\
    \midrule
    MVD-HPE (naive triangulation) & - & 95.4 & 92.6 & 95.2 & 93.7 & 94.2\\
    MVD-HPE (naive triangulation) & \checkmark & 95.9 & 92.8 & 96.0 & 94.1 & 94.7\\
    MVD-HPE ($w_{k,d} \equiv 1$ if exist) & - & 90.1 & 85.2 & 90.4 & 89.5 & 88.8\\
    MVD-HPE ($w_{k,d} \equiv 1$ if exist) & \checkmark & 90.2 & 85.6 & 90.9 & 90.3 & 89.3\\
    \midrule
     MVD-HPE (w/o depth) & - & 96.9 & 95.8 & 97.7 & 95.3 & 96.4\\
     MVD-HPE (w/o ICP) & - & 95.9 & 95.1 & 97.0 & 94.8 & 95.7\\
     MVD-HPE (w/o depth \& ICP) & - & 95.3 & 94.7 & 96.3 & 94.5 & 95.2\\
     \midrule
     MVD-HPE (Ours) w/ 5 cameras & - & 97.1 & 95.7 & 98.4 & 95.4 & 96.7\\
     MVD-HPE (Ours) w/ 4 cameras & - & 96.4 & 95.1 & 97.6 & 94.8 & 96.0\\
     MVD-HPE (Ours) w/ 3 cameras & - & 94.5 & 93.3 & 96.1 & 93.8 & 94.4\\
     MVD-HPE (Ours) w/ 2 cameras & - & 90.1 & 89.2 & 93.5 & 88.6 & 90.4\\
    \bottomrule
  \end{tabular}
\end{table*}
\subsection{Results}

\paragraph{Camera pose estimation.} We compare proposed MVD-HPE primarily with two baselines: Wide-Baseline~\cite{xu2021wide} and UncaliPose~\cite{Xu_2022_BMVC} for camera calibration. For completeness, we also compare MVD-HPE against two geometrical baselines: 1) use SIFT~\cite{lowe2004distinctive} feature or 2) SuperPoint~\cite{detone2018superpoint} as the descriptor, and match the key points across cameras using Brute-Force Matching (BFM)~\cite{jakubovic2018image}. Since we have 3D point clouds from each depth camera, we can use iterative closest point (ICP)~\cite{chetverikov2002trimmed} to improve all the mentioned baselines further. Table~\ref{tab:cam_pose_err} reports camera pose errors in both rotation and translation. Interestingly, we observe that the traditional baselines like SIFT~\cite{lowe2004distinctive} or SuperPoint~\cite{detone2018superpoint} perform poorly on sparse multi-depth cameras as the cross-view correspondences are extremely noisy due to limited 3D coverage and background clutter - this observation is consistent with \cite{xu2021wide}. Compared to Wide-Baseline~\cite{xu2021wide} and UncaliPose~\cite{Xu_2022_BMVC}, MVD-HPE w/o depth-guide, which leverages 3D re-ID features, exhibits superior performance. Implying that 3D representations are well suited for discriminative tasks like cross-view re-ID compared to image-only representations. Furthermore, our proposed full-system, i.e., MVD-HPE with depth-guided camera pose estimation, significantly improves the average error by $74$ mm and $0.83^{\circ}$ compared to the prior art. We also report the performance with ``idea matches'' in step B: cross-view matching to ablate the errors brought by clustering.

\begin{figure}[t!]
  \centering
  \includegraphics[width=\linewidth]{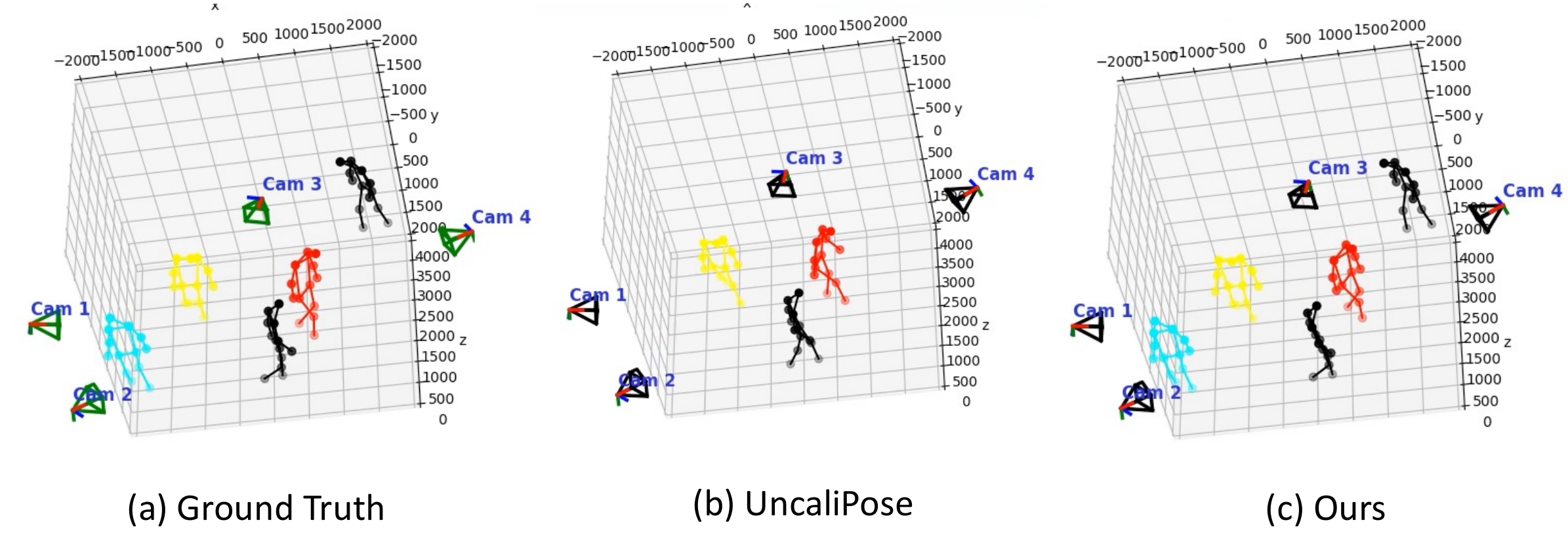}
  \vspace{-5mm}
  \caption{{Qualitative comparison of uncalibrated \textbf{3D human pose estimation} on the Garage video sets}. We compare our approach with UncaliPose~\cite{Xu_2022_BMVC}.}
  \label{fig:pose_garage_eval}
  \vspace{-4mm}
\end{figure}

\paragraph{3D human pose estimation.} 
We compare MVD-HPE against baselines using both calibrated (\cite{belagiannis20153d,ershadi2018multiple,dong2019fast}) and uncalibrated (\cite{Xu_2022_BMVC}) cameras in Table~\ref{tab:pcp_sota}. Compared to Dong \etal~\cite{dong2019fast}, Ershadi \etal~\cite{ershadi2018multiple}, and Belagiannis \etal~\cite{belagiannis20153d} which use calibrated cameras to predict 3D human poses, MVD-HPE with calibrated cameras produces superior results. This implies that our proposed depth constraint is able to reconstruct 3D poses even with sparse views accurately. Next, when cameras are uncalibrated, MVD-HPE still achieves state-of-the-art results compared to previous methods. Note the performance gap (around 1\%) between using calibrated versus uncalibrated cameras is indicative of errors induced in the camera pose estimation step. 
In addition, we also visualized one example of the comparison with UncaliPose~\cite{Xu_2022_BMVC} in Figure~\ref{fig:pose_garage_eval}. Even though some of the humans are outside the center capturing area, our approach still produces the correct 3D poses given sparse cameras.

\subsection{Ablation Studies}
In this section, we perform studies to highlight the contributions of each objective in our proposed method; refer to Table~\ref{tab:pcp_abl}.

\paragraph{Depth-constrained 3D human pose optimization.}
In our ablations, first, we evaluate variants of MVD-HPE with naive triangulation ("naive triangulation only") and using all depth projected 3D points ("$w_{k,d} \equiv 1 $") in the second section of Table~\ref{tab:pcp_abl}. We show that compared to naive triangulation, our proposed objective with depth regularization showcases significant improvement in performance. Importantly, since the depth measurements from RGBD cameras can be very noisy (see Figure~\ref{fig:constraint} (a)), simply using all the 3D projected points can result in inferior performance compared with naive triangulation.

\paragraph{Depth-guided pose estimation for 3D human pose.}
To assess the effectiveness of each component in our proposed depth-guided camera pose estimation, we ablate MVD-HPE with depth guidance and ICP excluded in the third section of Table~\ref{tab:pcp_abl}. Our results highlight the importance of using 3D depth constraints during human pose estimation from limited views. The 3D pose estimates are refined further by using registration by ICP. Additionally, we also conclude that incorrect camera poses result in a major drop in 3D human pose estimation performance.

\paragraph{Number of cameras.} To further assess the effectiveness of our proposed MVD-HPE for tackling the challenge of using fewer cameras, we conduct ablation studies on the number of cameras; the results are presented in the fourth section of Table~\ref{tab:pcp_abl}. We can observe that our model achieves comparable performance as UncaliPose~\cite{Xu_2022_BMVC} (which uses six cameras) with only three depth cameras. This demonstrates that leveraging few depths (RGBD), cameras can potentially generate comparable 3D human poses than using more RGB cameras.

\section{Conclusions}
We tackle the problem of multi-person 3D pose estimation from limited RGBD cameras. The key challenge in this is associating key points across camera views and localizing them in 3D. In this regard,
we proposed a regression-free method, MVD-HPE, to predict accurate 3D poses from sparse multi-view depth videos. We design MVD-HPE to work with uncalibrated cameras and generalize across 3D scenes. We utilize 3D Re-ID appearance features from RGBD images to obtain cross-view correspondences. We show that our proposed 1) depth-guided camera-pose estimation and 2) depth-constrained 3D human pose estimation using depth-projected 3D points effectively jointly predicts camera poses and 3D poses. We validate MVD-HPE under diverse conditions by constructing a dataset containing RGBD videos obtained from a few depth cameras. Our results show that MVD-HPE outperforms existing state-of-the-art 3D regression-free methods across various camera settings and human activities.

{
    \small
    \bibliographystyle{ieeenat_fullname}
    \bibliography{main}

\begin{thebibliography}{68}
\providecommand{\natexlab}[1]{#1}
\providecommand{\url}[1]{\texttt{#1}}
\expandafter\ifx\csname urlstyle\endcsname\relax
  \providecommand{\doi}[1]{doi: #1}\else
  \providecommand{\doi}{doi: \begingroup \urlstyle{rm}\Url}\fi

\bibitem[Balntas et~al.(2018)Balntas, Li, and Prisacariu]{balntas2018relocnet}
Vassileios Balntas, Shuda Li, and Victor Prisacariu.
\newblock Relocnet: Continuous metric learning relocalisation using neural nets.
\newblock In \emph{Proceedings of the European Conference on Computer Vision (ECCV)}, pages 751--767, 2018.

\bibitem[Belagiannis et~al.(2014)Belagiannis, Wang, Schiele, Fua, Ilic, and Navab]{belagiannis2014multiple}
Vasileios Belagiannis, Xinchao Wang, Bernt Schiele, Pascal Fua, Slobodan Ilic, and Nassir Navab.
\newblock Multiple human pose estimation with temporally consistent 3d pictorial structures.
\newblock In \emph{European Conference on Computer Vision}, pages 742--754. Springer, 2014.

\bibitem[Belagiannis et~al.(2015)Belagiannis, Amin, Andriluka, Schiele, Navab, and Ilic]{belagiannis20153d}
Vasileios Belagiannis, Sikandar Amin, Mykhaylo Andriluka, Bernt Schiele, Nassir Navab, and Slobodan Ilic.
\newblock 3d pictorial structures revisited: Multiple human pose estimation.
\newblock \emph{IEEE transactions on pattern analysis and machine intelligence}, 38\penalty0 (10):\penalty0 1929--1942, 2015.

\bibitem[Benbihi et~al.(2019)Benbihi, Geist, and Pradalier]{benbihi2019elf}
Assia Benbihi, Matthieu Geist, and Cedric Pradalier.
\newblock Elf: Embedded localisation of features in pre-trained cnn.
\newblock In \emph{Proceedings of the IEEE International Conference on Computer Vision}, pages 7940--7949, 2019.

\bibitem[Benzine et~al.(2020)Benzine, Chabot, Luvison, Pham, and Achard]{benzine2020pandanet}
Abdallah Benzine, Florian Chabot, Bertrand Luvison, Quoc~Cuong Pham, and Catherine Achard.
\newblock Pandanet: Anchor-based single-shot multi-person 3d pose estimation.
\newblock In \emph{Proceedings of the IEEE/CVF Conference on Computer Vision and Pattern Recognition}, pages 6856--6865, 2020.

\bibitem[Burenius et~al.(2013)Burenius, Sullivan, and Carlsson]{burenius20133d}
Magnus Burenius, Josephine Sullivan, and Stefan Carlsson.
\newblock 3d pictorial structures for multiple view articulated pose estimation.
\newblock In \emph{Proceedings of the IEEE conference on computer vision and pattern recognition}, pages 3618--3625, 2013.

\bibitem[Calonder et~al.(2010)Calonder, Lepetit, Strecha, and Fua]{calonder2010brief}
Michael Calonder, Vincent Lepetit, Christoph Strecha, and Pascal Fua.
\newblock Brief: Binary robust independent elementary features.
\newblock In \emph{European conference on computer vision}, pages 778--792. Springer, 2010.

\bibitem[Cao et~al.(2017)Cao, Simon, Wei, and Sheikh]{cao2017realtime}
Zhe Cao, Tomas Simon, Shih-En Wei, and Yaser Sheikh.
\newblock Realtime multi-person 2d pose estimation using part affinity fields.
\newblock In \emph{Proceedings of the IEEE conference on computer vision and pattern recognition}, pages 7291--7299, 2017.

\bibitem[Chen and Ramanan(2017)]{chen20173d}
Ching-Hang Chen and Deva Ramanan.
\newblock 3d human pose estimation= 2d pose estimation+ matching.
\newblock In \emph{Proceedings of the IEEE Conference on Computer Vision and Pattern Recognition}, pages 7035--7043, 2017.

\bibitem[Cheng et~al.(2020)Cheng, Xiao, Wang, Shi, Huang, and Zhang]{cheng2020higherhrnet}
Bowen Cheng, Bin Xiao, Jingdong Wang, Honghui Shi, Thomas~S Huang, and Lei Zhang.
\newblock Higherhrnet: Scale-aware representation learning for bottom-up human pose estimation.
\newblock In \emph{Proceedings of the IEEE/CVF conference on computer vision and pattern recognition}, pages 5386--5395, 2020.

\bibitem[Chetverikov et~al.(2002)Chetverikov, Svirko, Stepanov, and Krsek]{chetverikov2002trimmed}
Dmitry Chetverikov, Dmitry Svirko, Dmitry Stepanov, and Pavel Krsek.
\newblock The trimmed iterative closest point algorithm.
\newblock In \emph{2002 International Conference on Pattern Recognition}, pages 545--548. IEEE, 2002.

\bibitem[DeTone et~al.(2018)DeTone, Malisiewicz, and Rabinovich]{detone2018superpoint}
Daniel DeTone, Tomasz Malisiewicz, and Andrew Rabinovich.
\newblock Superpoint: Self-supervised interest point detection and description.
\newblock In \emph{Proceedings of the IEEE Conference on Computer Vision and Pattern Recognition Workshops}, pages 224--236, 2018.

\bibitem[Ding et~al.(2019)Ding, Wang, Sun, Shi, and Luo]{ding2019camnet}
Mingyu Ding, Zhe Wang, Jiankai Sun, Jianping Shi, and Ping Luo.
\newblock Camnet: Coarse-to-fine retrieval for camera re-localization.
\newblock In \emph{Proceedings of the IEEE International Conference on Computer Vision}, pages 2871--2880, 2019.

\bibitem[Dong et~al.(2019)Dong, Jiang, Huang, Bao, and Zhou]{dong2019fast}
Junting Dong, Wen Jiang, Qixing Huang, Hujun Bao, and Xiaowei Zhou.
\newblock Fast and robust multi-person 3d pose estimation from multiple views.
\newblock In \emph{Proceedings of the IEEE/CVF Conference on Computer Vision and Pattern Recognition}, pages 7792--7801, 2019.

\bibitem[Dusmanu et~al.(2019)Dusmanu, Rocco, Pajdla, Pollefeys, Sivic, Torii, and Sattler]{dusmanu2019d2}
Mihai Dusmanu, Ignacio Rocco, Tomas Pajdla, Marc Pollefeys, Josef Sivic, Akihiko Torii, and Torsten Sattler.
\newblock D2-net: A trainable cnn for joint description and detection of local features.
\newblock In \emph{Proceedings of the IEEE Conference on Computer Vision and Pattern Recognition}, pages 8092--8101, 2019.

\bibitem[Ershadi-Nasab et~al.(2018)Ershadi-Nasab, Noury, Kasaei, and Sanaei]{ershadi2018multiple}
Sara Ershadi-Nasab, Erfan Noury, Shohreh Kasaei, and Esmaeil Sanaei.
\newblock Multiple human 3d pose estimation from multiview images.
\newblock \emph{Multimedia Tools and Applications}, 77\penalty0 (12):\penalty0 15573--15601, 2018.

\bibitem[Fabbri et~al.(2020)Fabbri, Lanzi, Calderara, Alletto, and Cucchiara]{fabbri2020compressed}
Matteo Fabbri, Fabio Lanzi, Simone Calderara, Stefano Alletto, and Rita Cucchiara.
\newblock Compressed volumetric heatmaps for multi-person 3d pose estimation.
\newblock In \emph{Proceedings of the IEEE/CVF Conference on Computer Vision and Pattern Recognition}, pages 7204--7213, 2020.

\bibitem[Georgiev and Radulov(2014)]{georgiev2014practical}
Georgi~Hristov Georgiev and Vencislav~Dakov Radulov.
\newblock A practical method for decomposition of the essential matrix.
\newblock \emph{Applied mathematical sciences}, 8\penalty0 (176):\penalty0 8755--8770, 2014.

\bibitem[Gu et~al.(2018)Gu, Wang, Kuen, Ma, Shahroudy, Shuai, Liu, Wang, Wang, Cai, et~al.]{gu2018recent}
Jiuxiang Gu, Zhenhua Wang, Jason Kuen, Lianyang Ma, Amir Shahroudy, Bing Shuai, Ting Liu, Xingxing Wang, Gang Wang, Jianfei Cai, et~al.
\newblock Recent advances in convolutional neural networks.
\newblock \emph{Pattern Recognition}, 77:\penalty0 354--377, 2018.

\bibitem[Harris et~al.(1988)Harris, Stephens, et~al.]{harris1988combined}
Christopher~G Harris, Mike Stephens, et~al.
\newblock A combined corner and edge detector.
\newblock In \emph{Alvey vision conference}, pages 10--5244. Citeseer, 1988.

\bibitem[Hartley and Zisserman(2003)]{hartley2003multiple}
Richard Hartley and Andrew Zisserman.
\newblock \emph{Multiple view geometry in computer vision}.
\newblock Cambridge university press, 2003.

\bibitem[Hartley(1997)]{hartley1997defense}
Richard~I Hartley.
\newblock In defense of the eight-point algorithm.
\newblock \emph{IEEE Transactions on pattern analysis and machine intelligence}, 19\penalty0 (6):\penalty0 580--593, 1997.

\bibitem[Hartley and Sturm(1997)]{hartley1997triangulation}
Richard~I Hartley and Peter Sturm.
\newblock Triangulation.
\newblock \emph{Computer vision and image understanding}, 68\penalty0 (2):\penalty0 146--157, 1997.

\bibitem[Jakubovi{\'c} and Velagi{\'c}(2018)]{jakubovic2018image}
Amila Jakubovi{\'c} and Jasmin Velagi{\'c}.
\newblock Image feature matching and object detection using brute-force matchers.
\newblock In \emph{2018 International Symposium ELMAR}, pages 83--86. IEEE, 2018.

\bibitem[Kendall et~al.(2015)Kendall, Grimes, and Cipolla]{kendall2015posenet}
Alex Kendall, Matthew Grimes, and Roberto Cipolla.
\newblock Posenet: A convolutional network for real-time 6-dof camera relocalization.
\newblock In \emph{Proceedings of the IEEE international conference on computer vision}, pages 2938--2946, 2015.

\bibitem[Kipf and Welling(2016)]{kipf2016semi}
Thomas~N Kipf and Max Welling.
\newblock Semi-supervised classification with graph convolutional networks.
\newblock \emph{arXiv preprint arXiv:1609.02907}, 2016.

\bibitem[Laskar et~al.(2017)Laskar, Melekhov, Kalia, and Kannala]{laskar2017camera}
Zakaria Laskar, Iaroslav Melekhov, Surya Kalia, and Juho Kannala.
\newblock Camera relocalization by computing pairwise relative poses using convolutional neural network.
\newblock In \emph{Proceedings of the IEEE International Conference on Computer Vision Workshops}, pages 929--938, 2017.

\bibitem[Li and Lee(2019)]{li2019generating}
Chen Li and Gim~Hee Lee.
\newblock Generating multiple hypotheses for 3d human pose estimation with mixture density network.
\newblock In \emph{Proceedings of the IEEE/CVF Conference on Computer Vision and Pattern Recognition}, pages 9887--9895, 2019.

\bibitem[Li and Hartley(2006)]{li2006five}
Hongdong Li and Richard Hartley.
\newblock Five-point motion estimation made easy.
\newblock In \emph{18th International Conference on Pattern Recognition (ICPR'06)}, pages 630--633. IEEE, 2006.

\bibitem[Li et~al.(2015)Li, Zhang, and Chan]{li2015maximum}
Sijin Li, Weichen Zhang, and Antoni~B Chan.
\newblock Maximum-margin structured learning with deep networks for 3d human pose estimation.
\newblock In \emph{Proceedings of the IEEE international conference on computer vision}, pages 2848--2856, 2015.

\bibitem[Li et~al.(2019)Li, Wang, Yin, Peng, Du, Xiao, Yu, Lu, Wei, and Sun]{li2019rethinking}
Wenbo Li, Zhicheng Wang, Binyi Yin, Qixiang Peng, Yuming Du, Tianzi Xiao, Gang Yu, Hongtao Lu, Yichen Wei, and Jian Sun.
\newblock Rethinking on multi-stage networks for human pose estimation.
\newblock \emph{arXiv preprint arXiv:1901.00148}, 2019.

\bibitem[Li et~al.(2021)Li, Weng, Xu, and Kitani]{li2021visio}
Yu-Jhe Li, Xinshuo Weng, Yan Xu, and Kris~M Kitani.
\newblock Visio-temporal attention for multi-camera multi-target association.
\newblock In \emph{Proceedings of the IEEE/CVF International Conference on Computer Vision}, pages 9834--9844, 2021.

\bibitem[Lowe(2004)]{lowe2004distinctive}
David~G Lowe.
\newblock Distinctive image features from scale-invariant keypoints.
\newblock \emph{International journal of computer vision}, 60\penalty0 (2):\penalty0 91--110, 2004.

\bibitem[Martinez et~al.(2017)Martinez, Hossain, Romero, and Little]{martinez2017simple}
Julieta Martinez, Rayat Hossain, Javier Romero, and James~J Little.
\newblock A simple yet effective baseline for 3d human pose estimation.
\newblock In \emph{Proceedings of the IEEE international conference on computer vision}, pages 2640--2649, 2017.

\bibitem[Mehta et~al.(2018)Mehta, Sotnychenko, Mueller, Xu, Sridhar, Pons-Moll, and Theobalt]{mehta2018single}
Dushyant Mehta, Oleksandr Sotnychenko, Franziska Mueller, Weipeng Xu, Srinath Sridhar, Gerard Pons-Moll, and Christian Theobalt.
\newblock Single-shot multi-person 3d pose estimation from monocular rgb.
\newblock In \emph{2018 International Conference on 3D Vision (3DV)}, pages 120--130. IEEE, 2018.

\bibitem[Mehta et~al.(2020)Mehta, Sotnychenko, Mueller, Xu, Elgharib, Fua, Seidel, Rhodin, Pons-Moll, and Theobalt]{mehta2020xnect}
Dushyant Mehta, Oleksandr Sotnychenko, Franziska Mueller, Weipeng Xu, Mohamed Elgharib, Pascal Fua, Hans-Peter Seidel, Helge Rhodin, Gerard Pons-Moll, and Christian Theobalt.
\newblock Xnect: Real-time multi-person 3d motion capture with a single rgb camera.
\newblock \emph{Acm Transactions On Graphics (TOG)}, 39\penalty0 (4):\penalty0 82--1, 2020.

\bibitem[Melekhov et~al.(2017)Melekhov, Ylioinas, Kannala, and Rahtu]{melekhov2017relative}
Iaroslav Melekhov, Juha Ylioinas, Juho Kannala, and Esa Rahtu.
\newblock Relative camera pose estimation using convolutional neural networks.
\newblock In \emph{International Conference on Advanced Concepts for Intelligent Vision Systems}, pages 675--687. Springer, 2017.

\bibitem[Moon et~al.(2019)Moon, Chang, and Lee]{moon2019camera}
Gyeongsik Moon, Ju~Yong Chang, and Kyoung~Mu Lee.
\newblock Camera distance-aware top-down approach for 3d multi-person pose estimation from a single rgb image.
\newblock In \emph{Proceedings of the ieee/cvf international conference on computer vision}, pages 10133--10142, 2019.

\bibitem[Moon(1996)]{moon1996expectation}
Todd~K Moon.
\newblock The expectation-maximization algorithm.
\newblock \emph{IEEE Signal processing magazine}, 13\penalty0 (6):\penalty0 47--60, 1996.

\bibitem[Muja and Lowe(2009)]{muja2009flann}
Marius Muja and David Lowe.
\newblock Flann-fast library for approximate nearest neighbors user manual.
\newblock \emph{Computer Science Department, University of British Columbia, Vancouver, BC, Canada}, 2009.

\bibitem[Nie et~al.(2019)Nie, Feng, Zhang, and Yan]{nie2019single}
Xuecheng Nie, Jiashi Feng, Jianfeng Zhang, and Shuicheng Yan.
\newblock Single-stage multi-person pose machines.
\newblock In \emph{Proceedings of the IEEE/CVF international conference on computer vision}, pages 6951--6960, 2019.

\bibitem[Nist{\'e}r(2004)]{nister2004efficient}
David Nist{\'e}r.
\newblock An efficient solution to the five-point relative pose problem.
\newblock \emph{IEEE transactions on pattern analysis and machine intelligence}, 26\penalty0 (6):\penalty0 756--770, 2004.

\bibitem[Olson(2011)]{olson2011apriltag}
Edwin Olson.
\newblock Apriltag: A robust and flexible visual fiducial system.
\newblock In \emph{2011 IEEE international conference on robotics and automation}, pages 3400--3407. IEEE, 2011.

\bibitem[Ono et~al.(2018)Ono, Trulls, Fua, and Yi]{ono2018lf}
Yuki Ono, Eduard Trulls, Pascal Fua, and Kwang~Moo Yi.
\newblock Lf-net: learning local features from images.
\newblock In \emph{Advances in neural information processing systems}, pages 6234--6244, 2018.

\bibitem[Pavlakos et~al.(2017)Pavlakos, Zhou, Derpanis, and Daniilidis]{pavlakos2017coarse}
Georgios Pavlakos, Xiaowei Zhou, Konstantinos~G Derpanis, and Kostas Daniilidis.
\newblock Coarse-to-fine volumetric prediction for single-image 3d human pose.
\newblock In \emph{Proceedings of the IEEE conference on computer vision and pattern recognition}, pages 7025--7034, 2017.

\bibitem[Pavlakos et~al.(2018)Pavlakos, Zhou, and Daniilidis]{pavlakos2018ordinal}
Georgios Pavlakos, Xiaowei Zhou, and Kostas Daniilidis.
\newblock Ordinal depth supervision for 3d human pose estimation.
\newblock In \emph{Proceedings of the IEEE Conference on Computer Vision and Pattern Recognition}, pages 7307--7316, 2018.

\bibitem[Reddy et~al.(2021)Reddy, Guigues, Pishchulin, Eledath, and Narasimhan]{reddy2021tessetrack}
N~Dinesh Reddy, Laurent Guigues, Leonid Pishchulin, Jayan Eledath, and Srinivasa~G Narasimhan.
\newblock Tessetrack: End-to-end learnable multi-person articulated 3d pose tracking.
\newblock In \emph{Proceedings of the IEEE/CVF Conference on Computer Vision and Pattern Recognition}, pages 15190--15200, 2021.

\bibitem[Redmon and Farhadi(2018)]{redmon2018yolov3}
Joseph Redmon and Ali Farhadi.
\newblock Yolov3: An incremental improvement.
\newblock \emph{arXiv preprint arXiv:1804.02767}, 2018.

\bibitem[Rogez et~al.(2017)Rogez, Weinzaepfel, and Schmid]{rogez2017lcr}
Gregory Rogez, Philippe Weinzaepfel, and Cordelia Schmid.
\newblock Lcr-net: Localization-classification-regression for human pose.
\newblock In \emph{Proceedings of the IEEE Conference on Computer Vision and Pattern Recognition}, pages 3433--3441, 2017.

\bibitem[Rogez et~al.(2019)Rogez, Weinzaepfel, and Schmid]{rogez2019lcr}
Gregory Rogez, Philippe Weinzaepfel, and Cordelia Schmid.
\newblock Lcr-net++: Multi-person 2d and 3d pose detection in natural images.
\newblock \emph{IEEE transactions on pattern analysis and machine intelligence}, 42\penalty0 (5):\penalty0 1146--1161, 2019.

\bibitem[Rosten and Drummond(2006)]{rosten2006machine}
Edward Rosten and Tom Drummond.
\newblock Machine learning for high-speed corner detection.
\newblock In \emph{European conference on computer vision}, pages 430--443. Springer, 2006.

\bibitem[Rublee et~al.(2011)Rublee, Rabaud, Konolige, and Bradski]{rublee2011orb}
Ethan Rublee, Vincent Rabaud, Kurt Konolige, and Gary Bradski.
\newblock Orb: An efficient alternative to sift or surf.
\newblock In \emph{2011 International conference on computer vision}, pages 2564--2571. Ieee, 2011.

\bibitem[Shavit and Ferens(2019)]{shavit2019introduction}
Yoli Shavit and Ron Ferens.
\newblock Introduction to camera pose estimation with deep learning.
\newblock \emph{arXiv preprint arXiv:1907.05272}, 2019.

\bibitem[Sun et~al.(2019)Sun, Xiao, Liu, and Wang]{sun2019deep}
Ke Sun, Bin Xiao, Dong Liu, and Jingdong Wang.
\newblock Deep high-resolution representation learning for human pose estimation.
\newblock In \emph{Proceedings of the IEEE/CVF Conference on Computer Vision and Pattern Recognition}, pages 5693--5703, 2019.

\bibitem[Triggs et~al.(1999)Triggs, McLauchlan, Hartley, and Fitzgibbon]{triggs1999bundle}
Bill Triggs, Philip~F McLauchlan, Richard~I Hartley, and Andrew~W Fitzgibbon.
\newblock Bundle adjustment—a modern synthesis.
\newblock In \emph{International workshop on vision algorithms}, pages 298--372. Springer, 1999.

\bibitem[Tu et~al.(2020)Tu, Wang, and Zeng]{tu2020voxelpose}
Hanyue Tu, Chunyu Wang, and Wenjun Zeng.
\newblock Voxelpose: Towards multi-camera 3d human pose estimation in wild environment.
\newblock In \emph{European Conference on Computer Vision}, pages 197--212. Springer, 2020.

\bibitem[Vaswani et~al.(2017)Vaswani, Shazeer, Parmar, Uszkoreit, Jones, Gomez, Kaiser, and Polosukhin]{vaswani2017attention}
Ashish Vaswani, Noam Shazeer, Niki Parmar, Jakob Uszkoreit, Llion Jones, Aidan~N Gomez, Lukasz Kaiser, and Illia Polosukhin.
\newblock Attention is all you need.
\newblock \emph{arXiv preprint arXiv:1706.03762}, 2017.

\bibitem[Wu et~al.(2021)Wu, Jin, Liu, Bai, Qian, Liu, and Ouyang]{wu2021graph}
Size Wu, Sheng Jin, Wentao Liu, Lei Bai, Chen Qian, Dong Liu, and Wanli Ouyang.
\newblock Graph-based 3d multi-person pose estimation using multi-view images.
\newblock In \emph{Proceedings of the IEEE/CVF International Conference on Computer Vision}, pages 11148--11157, 2021.

\bibitem[Xu(2022)]{xu2022joint}
Yan Xu.
\newblock \emph{Joint Reasoning for Camera and 3D Human Pose Estimation}.
\newblock PhD thesis, Carnegie Mellon University, 2022.

\bibitem[Xu and Kitani(2022)]{Xu_2022_BMVC}
Yan Xu and Kris Kitani.
\newblock Multi-view multi-person 3d pose estimation with uncalibrated camera networks.
\newblock In \emph{33rd British Machine Vision Conference 2022, {BMVC} 2022, London, UK, November 21-24, 2022}. {BMVA} Press, 2022.

\bibitem[Xu et~al.(2020)Xu, Roy, and Kitani]{xu2020estimate}
Yan Xu, Vivek Roy, and Kris Kitani.
\newblock Estimate 3d camera pose from 2d pedestrian trajectories.
\newblock In \emph{The IEEE Winter Conference on Applications of Computer Vision}, pages 2579--2588, 2020.

\bibitem[Xu et~al.(2021)Xu, Li, Weng, and Kitani]{xu2021wide}
Yan Xu, Yu-Jhe Li, Xinshuo Weng, and Kris Kitani.
\newblock Wide-baseline multi-camera calibration using person re-identification.
\newblock In \emph{Proceedings of the IEEE/CVF Conference on Computer Vision and Pattern Recognition}, pages 13134--13143, 2021.

\bibitem[Yang et~al.(2020)Yang, Nguyen, Heijnen, and Balntas]{yang2020ur2kid}
Tsun-Yi Yang, Duy-Kien Nguyen, Huub Heijnen, and Vassileios Balntas.
\newblock Ur2kid: Unifying retrieval, keypoint detection, and keypoint description without local correspondence supervision.
\newblock \emph{arXiv preprint arXiv:2001.07252}, 2020.

\bibitem[Yi et~al.(2016)Yi, Trulls, Lepetit, and Fua]{yi2016lift}
Kwang~Moo Yi, Eduard Trulls, Vincent Lepetit, and Pascal Fua.
\newblock Lift: Learned invariant feature transform.
\newblock In \emph{European Conference on Computer Vision}, pages 467--483. Springer, 2016.

\bibitem[Zanfir et~al.(2018{\natexlab{a}})Zanfir, Marinoiu, and Sminchisescu]{zanfir2018monocular}
Andrei Zanfir, Elisabeta Marinoiu, and Cristian Sminchisescu.
\newblock Monocular 3d pose and shape estimation of multiple people in natural scenes-the importance of multiple scene constraints.
\newblock In \emph{Proceedings of the IEEE Conference on Computer Vision and Pattern Recognition}, pages 2148--2157, 2018{\natexlab{a}}.

\bibitem[Zanfir et~al.(2018{\natexlab{b}})Zanfir, Marinoiu, Zanfir, Popa, and Sminchisescu]{zanfir2018deep}
Andrei Zanfir, Elisabeta Marinoiu, Mihai Zanfir, Alin-Ionut Popa, and Cristian Sminchisescu.
\newblock Deep network for the integrated 3d sensing of multiple people in natural images.
\newblock \emph{Advances in Neural Information Processing Systems}, 31, 2018{\natexlab{b}}.

\bibitem[Zhang et~al.(2021)Zhang, Cai, Yan, Feng, et~al.]{zhang2021direct}
Jianfeng Zhang, Yujun Cai, Shuicheng Yan, Jiashi Feng, et~al.
\newblock Direct multi-view multi-person 3d pose estimation.
\newblock \emph{Advances in Neural Information Processing Systems}, 34, 2021.

\bibitem[Zheng et~al.(2022)Zheng, Wang, Zheng, and Yang]{zheng2022parameter}
Zhedong Zheng, Xiaohan Wang, Nenggan Zheng, and Yi Yang.
\newblock Parameter-efficient person re-identification in the 3d space.
\newblock \emph{IEEE Transactions on Neural Networks and Learning Systems}, 2022.

\end{thebibliography}
}

\clearpage
\appendix
\section{Appendix}
\subsection{Theory of projected 3D points from depths}\label{theory:trans}
Here we provide more details regarding the basic geometry of projecting 2D body key points to 3D with depth. We first denote the intrinsics $K_c$ and $K_d$ for the RGB camera and depth camera, respectively. We can also denote the input RGB image as $I_{rgb} \in \mathbb{R}^{h_c\times w_c \times 3}$ and the depth image as $I_{d} \in \mathbb{R}^{h_d\times w_d \times 1}$. For each 2D pixel ($x_d,y_d$) in $I_{d}$ we can derive the coordinate of point cloud in 3D as ($X_d,Y_d,Z_d$):

\begin{equation}
\begin{split}
\begin{bmatrix}
X_d\\
Y_d\\
Z_d
\end{bmatrix} = &  I_d{(x,y)} \cdot K_d^{-1} \cdot \begin{bmatrix}
x_d\\
y_d\\
1
\end{bmatrix}
\\
= &
I_d{(x_d,y_d)} \cdot \begin{bmatrix}
1/f_{xd} & 0 & -c_{xd}/f_{xd}\\
0 & 1/f_{yd} & -c_{yd}/f_{yd}\\
0 & 0 & 1
\end{bmatrix} \cdot \begin{bmatrix}
x_d\\
y_d\\
1
\end{bmatrix}
\end{split}
\end{equation},
where $f$ indicates the focal length and the $c$ indicates camera center.
However, the detected 2D key points with the human pose detectors are in the RGB image $I_{rgb}$. Hence, we will need to generate the transformed depth image  $I_{d}' \in \mathbb{R}^{h_c\times w_c \times 1}$ as the same size of height and width as $I_{rgb}$:
$$
\begin{bmatrix}
x_c\\
y_c\\
1
\end{bmatrix}
=\alpha \cdot K_c \cdot
\begin{bmatrix}
R_{dc} & T_{dc}
\end{bmatrix}
\begin{bmatrix}
X_d\\
Y_d\\
Z_d\\
1
\end{bmatrix}
=\alpha \cdot K_c \cdot
\begin{bmatrix}
X_c\\
Y_c\\
Z_c
\end{bmatrix}
$$,
where $R_{dc}$ and $T_{dc}$ indicate the transformation between depth camera and rgb camera, and $\alpha$ indicates the scale for the transformation to the 2D projective vector space.
We can then have $I_{d}'(x_c,y_c) = Z_c$ if there is a depth pixel mapped to the RGB pixel. Otherwise, $I_{d}'(x_c,y_c) = 0$ indicates no measure depth points. Hence, we can get the projected 3D points $P \in \mathbb{R}^{J \times 3}$ from each 2D skeleton $p \in \mathbb{R}^{J \times 2}$ using:

\begin{equation}
\begin{split}
\begin{bmatrix}
X_c\\
Y_c\\
Z_c
\end{bmatrix} & = I_d'{(x_c,y_c)} \cdot K_c^{-1} \cdot \begin{bmatrix}
x_c\\
y_c\\
1
\end{bmatrix}
\\ & =
I_d'{(x_d,y_d)} \cdot \begin{bmatrix}
1/f_{xc} & 0 & -c_x/f_{xc}\\
0 & 1/f_{yc} & -c_y/f_{yc}\\
0 & 0 & 1
\end{bmatrix} \cdot \begin{bmatrix}
x_c\\
y_c\\
1
\end{bmatrix}
\end{split}
\end{equation}.

\subsection{Implementation Details}

\paragraph{Datasets.} All of the video sets (office, garage, and classroom) are recorded by Microsoft Azure Kinect devices with audio cables plugged in to make them synchronized. The resolution of RGB cameras is $1280\times720$ (720p) while the resolution of depth cameras is $576\times480$. The Kinect provides distortion factors as well as the intrinsics. Hence, we undistort all of the RGB and depth images in advance. The depth meter uses millimeter for the depth maps recorded with Kinect thus we use the millimeter as the scale in the experiments.

\paragraph{Pipeline.}  We first detect 2D body key points across all camera views using YOLOv3~\cite{redmon2018yolov3} for bounding box detection and HRNet~\cite{sun2019deep} for top-down pose estimation. We utilize the pre-trained weights provided by HRNet\cite{sun2019deep}: "pose\_hrnet\_w48\_384x288.pth" to obtain the 2D poses. While there are $J=17$ key points in the 2D poses, we removed the four key points including ears and noses for simplicity and ended up with $J=13$ key points. The cropped RGBD format is obtained using the theory in Sec.~\ref{theory:trans} by first obtaining the transformed depth image, which is the same size as the RGB image. The RGBD image can be transformed into colored point clouds ($(x,y,z,r,g,b)$) using the same theory in Sec.~\ref{theory:trans}. We can then use the pre-trained OG-Net~\cite{zheng2022parameter} to derive the feature with dimension $512$ for each cropped RGBD image. The equation~\ref{eq:clustering} employs constrained K-Means to resolve. equation~\ref{eq:camera_pose} and equation~\ref{eq:triangulation} employ gradient descent to minimize the objective while grid-searching for the best $w$ in both of the equations. We use the threshold of $0.01$ and $10$ for the objective of equation~\ref{eq:camera_pose} and equation~\ref{eq:triangulation} respectively for searching the maximum $\sum w$. All of the experiments are run on a single Nvidia 2080 Ti GPU

\subsection{More qualitative results}
We now present more qualitative results and comparisons here due to the limited space in the main paper.

\begin{figure*}[t!]
  \centering
  \includegraphics[width=\linewidth]{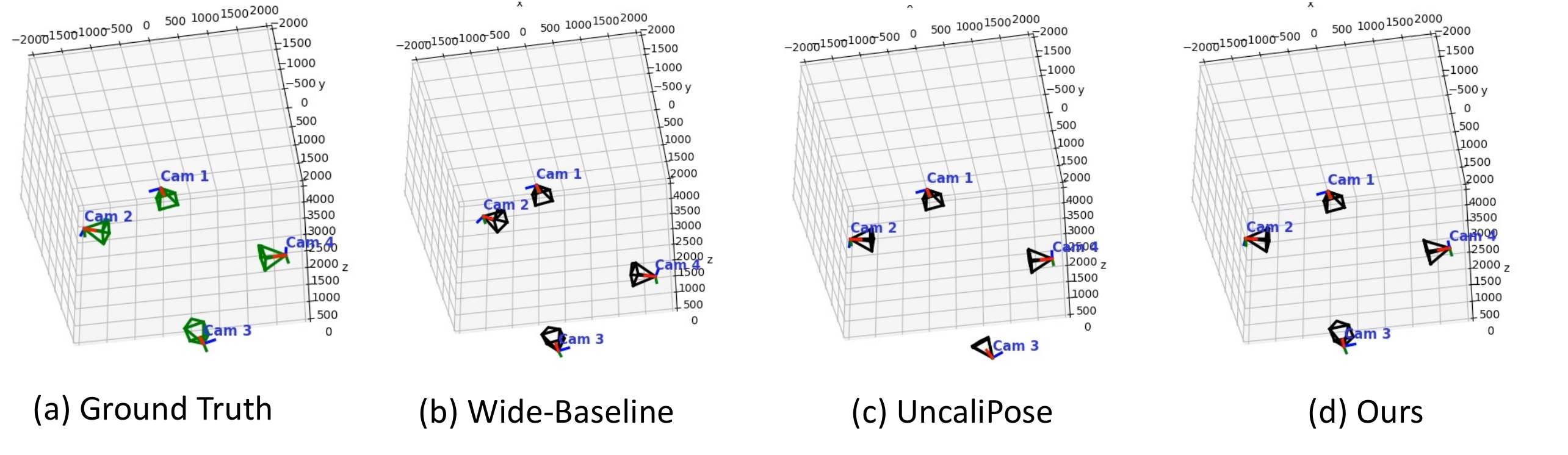}
  \vspace{-5mm}
  \caption{Qualitative comparison of \textbf{camera pose estimation} on the Office video sets. We compare our approach with Wide-Baseline~\cite{xu2021wide} and UncaliPose~\cite{Xu_2022_BMVC}.}
  \label{fig:camera_office}
  \vspace{-4mm}
\end{figure*}

\begin{figure*}[t!]
  \centering
  \includegraphics[width=\linewidth]{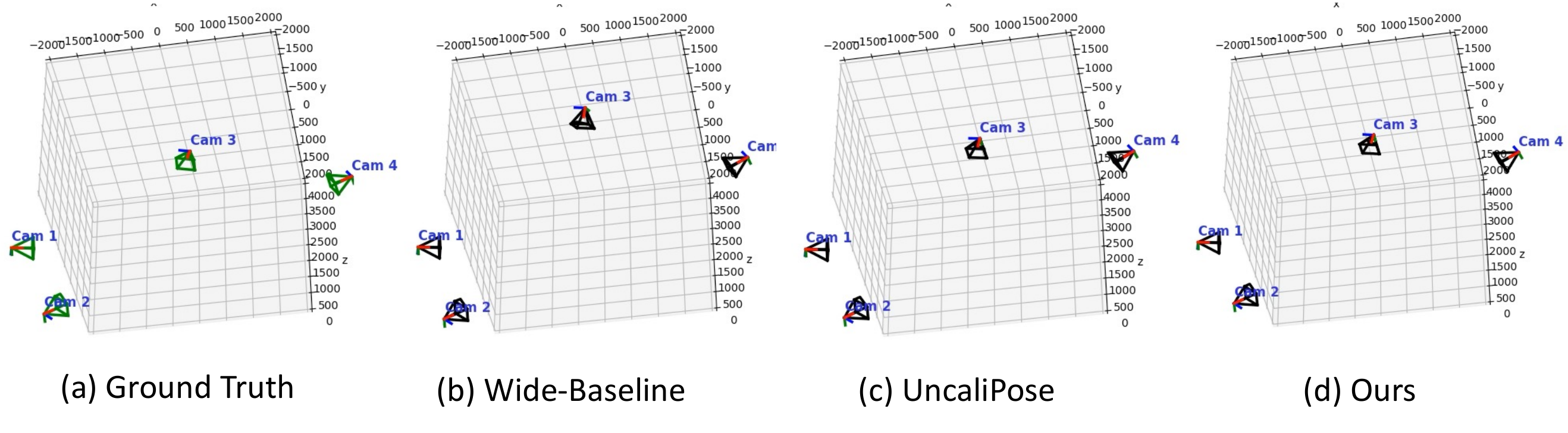}
  \vspace{-5mm}
  \caption{{Qualitative comparison of \textbf{camera pose estimation} on the Garage video sets}. We compare our approach with Wide-Baseline~\cite{xu2021wide} and UncaliPose~\cite{Xu_2022_BMVC}.}
  \label{fig:camera_garage}
  \vspace{-4mm}
\end{figure*}

\begin{figure*}[t!]
  \centering
  \includegraphics[width=\linewidth]{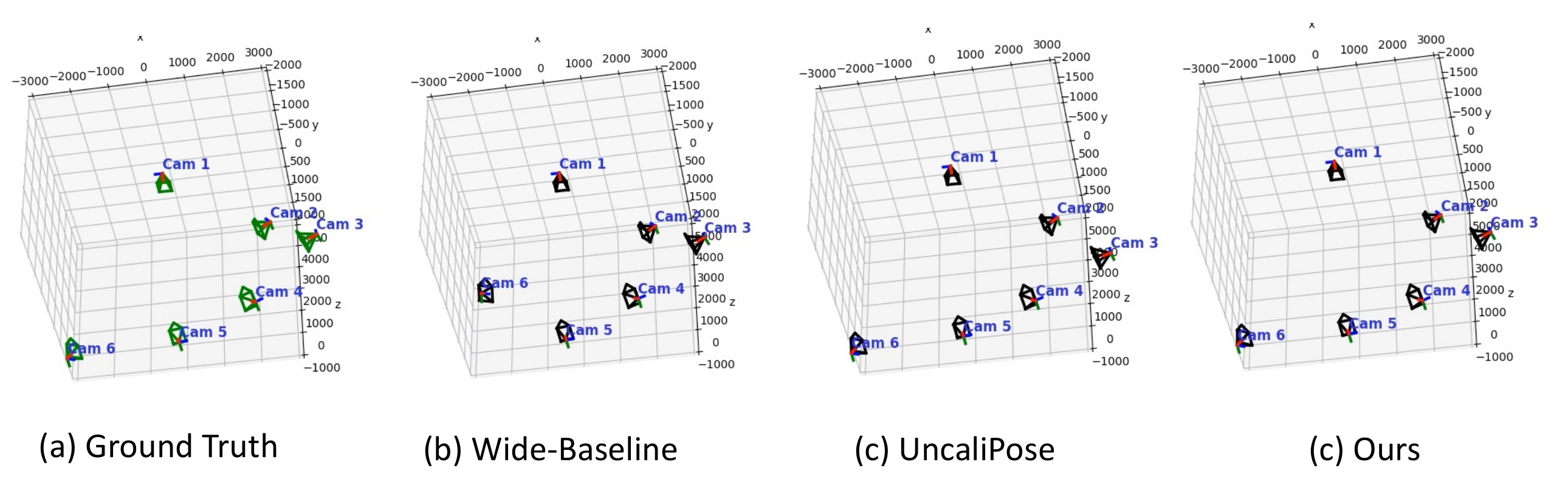}
  \vspace{-5mm}
  \caption{{Qualitative comparison of \textbf{camera pose estimation} on the Classroom video sets}. We compare our approach with Wide-Baseline~\cite{xu2021wide} and UncaliPose~\cite{Xu_2022_BMVC}.}
  \label{fig:camera_class}
  \vspace{-4mm}
\end{figure*}

\paragraph{Camera pose estimation.}
We compare our approach with Wide-Baseline~\cite{xu2021wide} and UncaliPose~\cite{Xu_2022_BMVC} and present the results in Figure~\ref{fig:camera_office}, Figure~\ref{fig:camera_garage}, and Figure~\ref{fig:camera_class}. We can observe that our approach achieves the best alignment with the ground truth camera poses. Compared with UncaliPose~\cite{Xu_2022_BMVC} which achieves the comparable with RGB-only feature, our model leveraging 3D features and depth information produces closer camera poses.


\begin{figure*}[t!]
  \centering
  \includegraphics[width=0.8\linewidth]{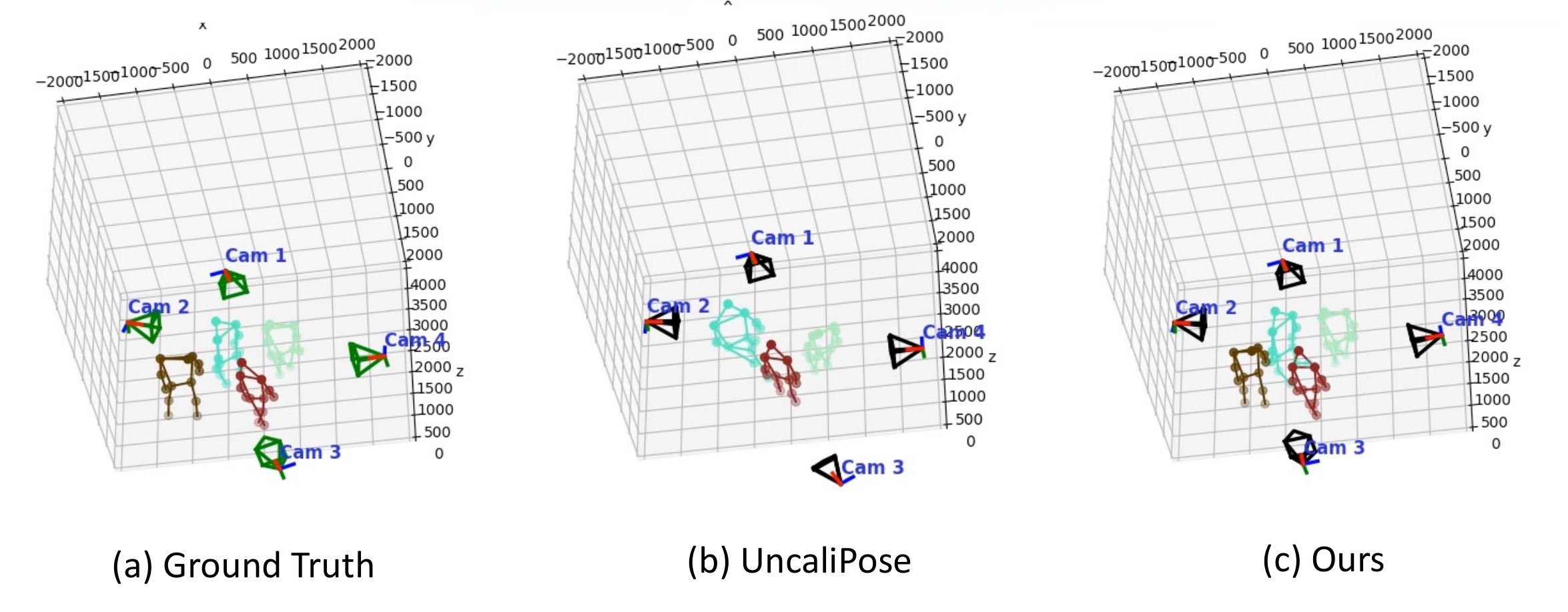}
  \vspace{-5mm}
  \caption{{Qualitative comparison of uncalibrated \textbf{3D human pose estimation} on the Office video sets}. We compare our approach with UncaliPose~\cite{Xu_2022_BMVC}.}
  \label{fig:pose_office}
  \vspace{-4mm}
\end{figure*}

\begin{figure*}[t!]
  \centering
  \includegraphics[width=0.8\linewidth]{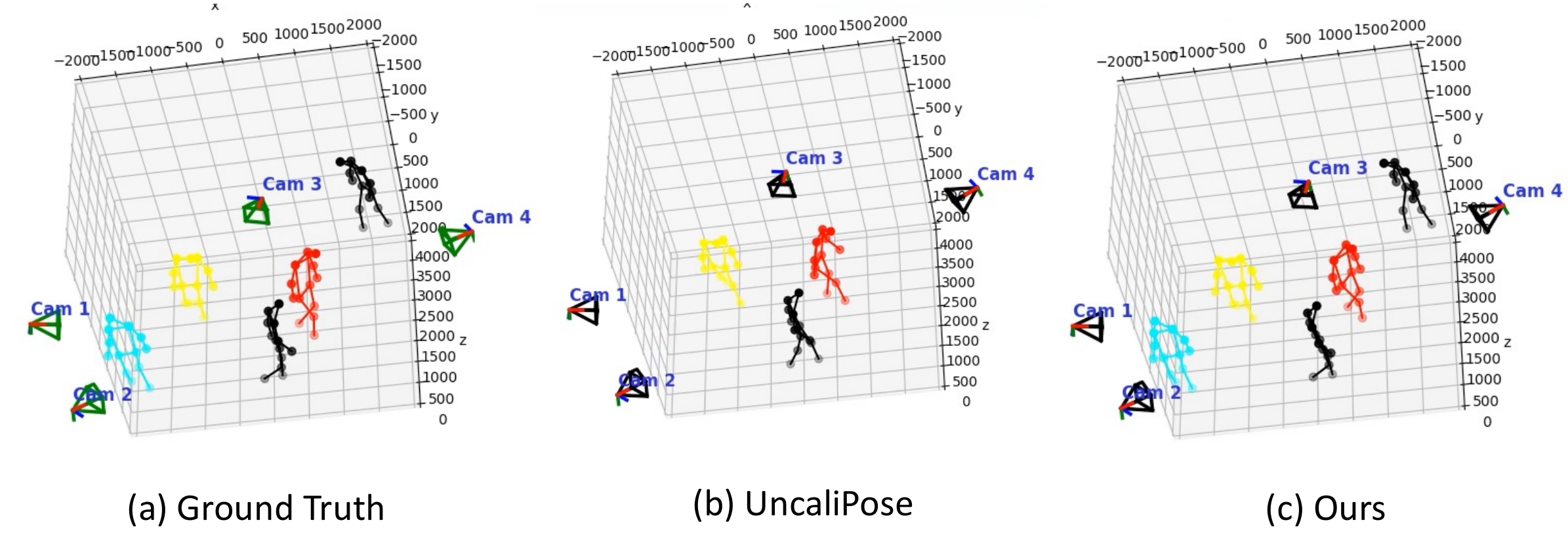}
  \vspace{-5mm}
  \caption{{Qualitative comparison of uncalibrated \textbf{3D human pose estimation} on the Garage video sets}. We compare our approach with UncaliPose~\cite{Xu_2022_BMVC}.}
  \label{fig:pose_garage}
  \vspace{-4mm}
\end{figure*}

\begin{figure*}[t!]
  \centering
  \includegraphics[width=0.8\linewidth]{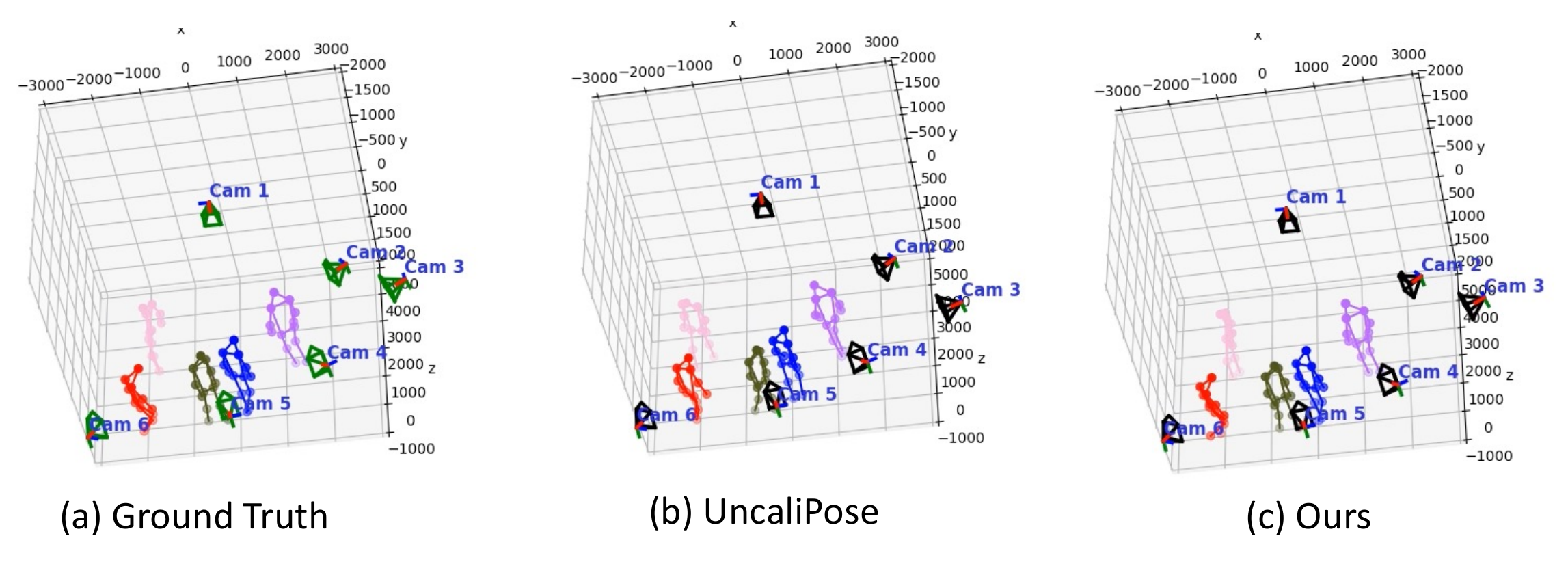}
  \vspace{-5mm}
  \caption{{Qualitative comparison of uncalibrated \textbf{3D human pose estimation} on the Classroom video sets}. We compare our approach with UncaliPose~\cite{Xu_2022_BMVC}.}
  \label{fig:pose_class}
  \vspace{-4mm}
\end{figure*}

\paragraph{3D Human pose estimation.}
We present the qualitative results of uncalibrated 3D human pose estimation and compare with UncaliPose~\cite{Xu_2022_BMVC} in Figure~\ref{fig:pose_office}, Figure~\ref{fig:pose_garage}, and Figure~\ref{fig:pose_class}. We can observe that even though UncaliPose~\cite{Xu_2022_BMVC} achieves comparable camera pose predictions to our method, the reconstructed 3D skeletons are still inferior due to lacking 3D guidance with depth information.


\subsection{Broader Impacts}

The emergence and advancement of multi-view 3D human pose estimation technology have had a profound social impact across various domains. One notable area is in healthcare, where this technology has revolutionized motion analysis and biomechanics research. By combining data from multiple camera views, multi-view 3D pose estimation systems can reconstruct and analyze human movements in three dimensions with higher accuracy and precision. This has facilitated more comprehensive assessments of gait patterns, posture, and joint movements, leading to improved diagnosis and treatment planning for conditions such as musculoskeletal disorders and neurological impairments. Moreover, in the field of sports and athletics, multi-view 3D human pose estimation has transformed training and performance analysis. Coaches and athletes can capture and synchronize movements from different angles, providing a holistic view of technique, form, and coordination. This allows for targeted interventions, personalized training programs, and objective performance evaluations, ultimately enhancing athletic performance and reducing the risk of injuries. Furthermore, the application of multi-view 3D human pose estimation extends to the field of computer vision and augmented reality. It enables realistic virtual avatars, immersive telepresence, and interactive virtual experiences, enhancing communication, entertainment, and creative expression. Overall, multi-view 3D human pose estimation has made significant contributions to healthcare, sports, computer vision, and entertainment, positively impacting individuals' lives, promoting innovation, and pushing the boundaries of human movement analysis. We believe our approach leveraging depth or Lidar (in the future) can have more impactful to these societies.

\subsection{Limitations}

While multi-view 3D pose estimation with depth cameras offers numerous advantages, it also comes with certain limitations. One key limitation is the reliance on line-of-sight visibility between the depth cameras and the subject being tracked. Obstacles such as furniture, walls, or other people can obstruct the view and result in occlusions, leading to incomplete or inaccurate pose estimations. Additionally, the performance of depth cameras can be affected by environmental factors such as lighting conditions, reflections, or interference from other devices operating in the same frequency range. These factors can introduce noise and artifacts in the depth data, compromising the accuracy of the pose estimation. Furthermore, the setup and calibration of multiple depth cameras for multi-view capture can be complex and time-consuming, requiring careful alignment and synchronization. Any errors in the calibration process can adversely impact the accuracy of the resulting 3D pose estimations. Another limitation lies in the computational demands of processing data from multiple-depth cameras in real time. The fusion and synchronization of data from multiple views require substantial computational resources, making it challenging to achieve real-time performance in certain scenarios. Despite these limitations, ongoing advancements in depth sensing technology, computer vision algorithms, and calibration techniques hold the promise of mitigating these challenges and improving the accuracy and usability of multi-view 3D pose estimation with depth cameras.

\subsection{Dataset regarding human subjects}

We follow the rules of IRB and have the consent (signed by them) from the participants in the data collection for each of the video sets:
\paragraph{Procedures.} 
The data collection procedure involves one or multiple participants in a data capture. This will take place in our conference room or our private office. You (the participant) will be asked to perform a task like walking, sitting, and standing within the capture zone. We will record the movements and body poses with multiple third-person depth cameras. THESE VIDEOS WILL CONTAIN IDENTIFIABLE FEATURES OF YOU (the participant). The data collection will be approximately 30 minutes.
\paragraph{Participant Requirements.} 
Participants must be at least 18 and need to be able to walk, be able to stand, and be able to sit for 30 mins.

More details of the consent form will be provided as supplementary files upon the camera-ready submission.

\newpage

\begin{figure*}[t!]
  \centering
  \includegraphics[width=\linewidth]{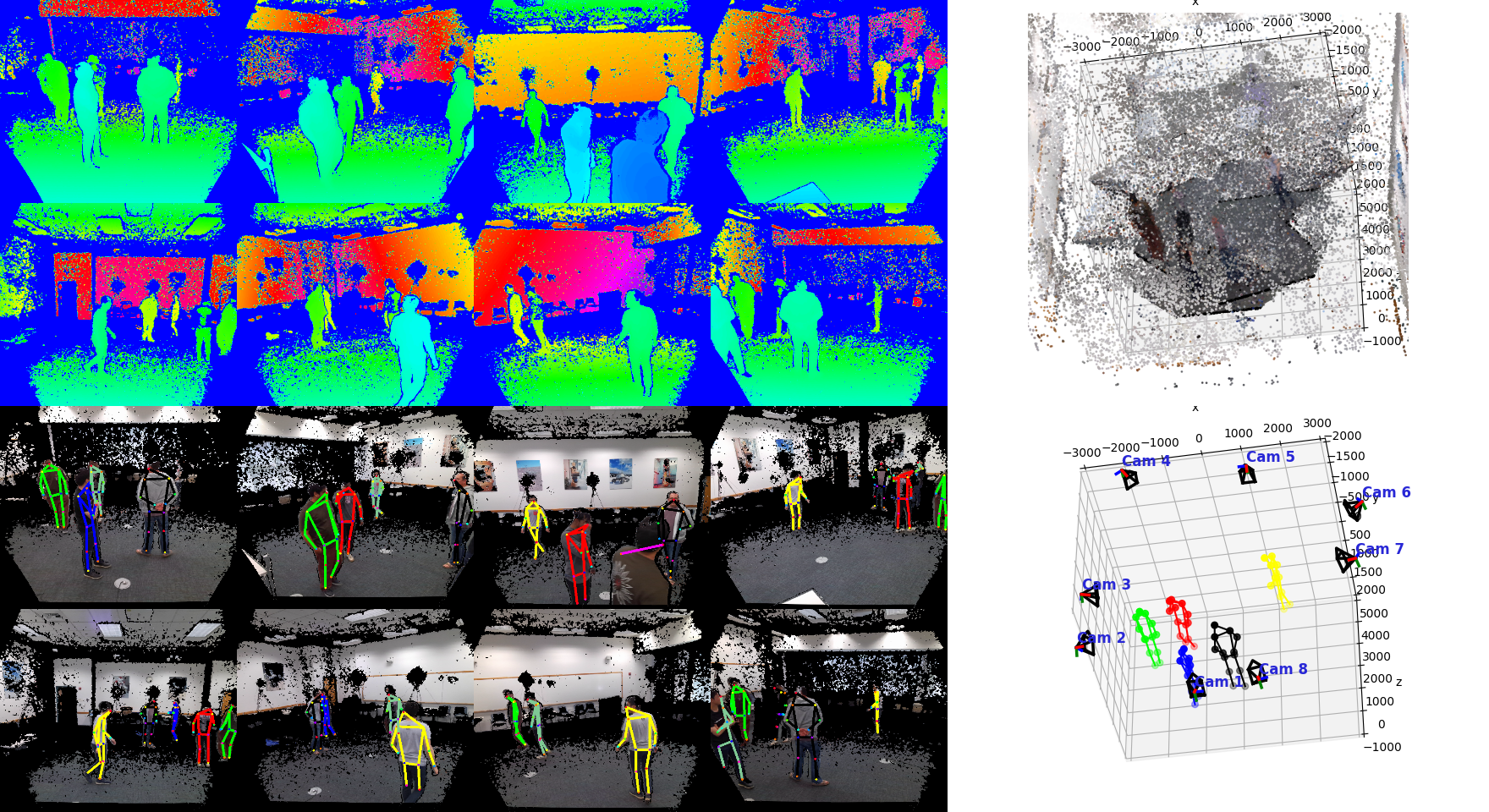}
  \includegraphics[width=\linewidth]{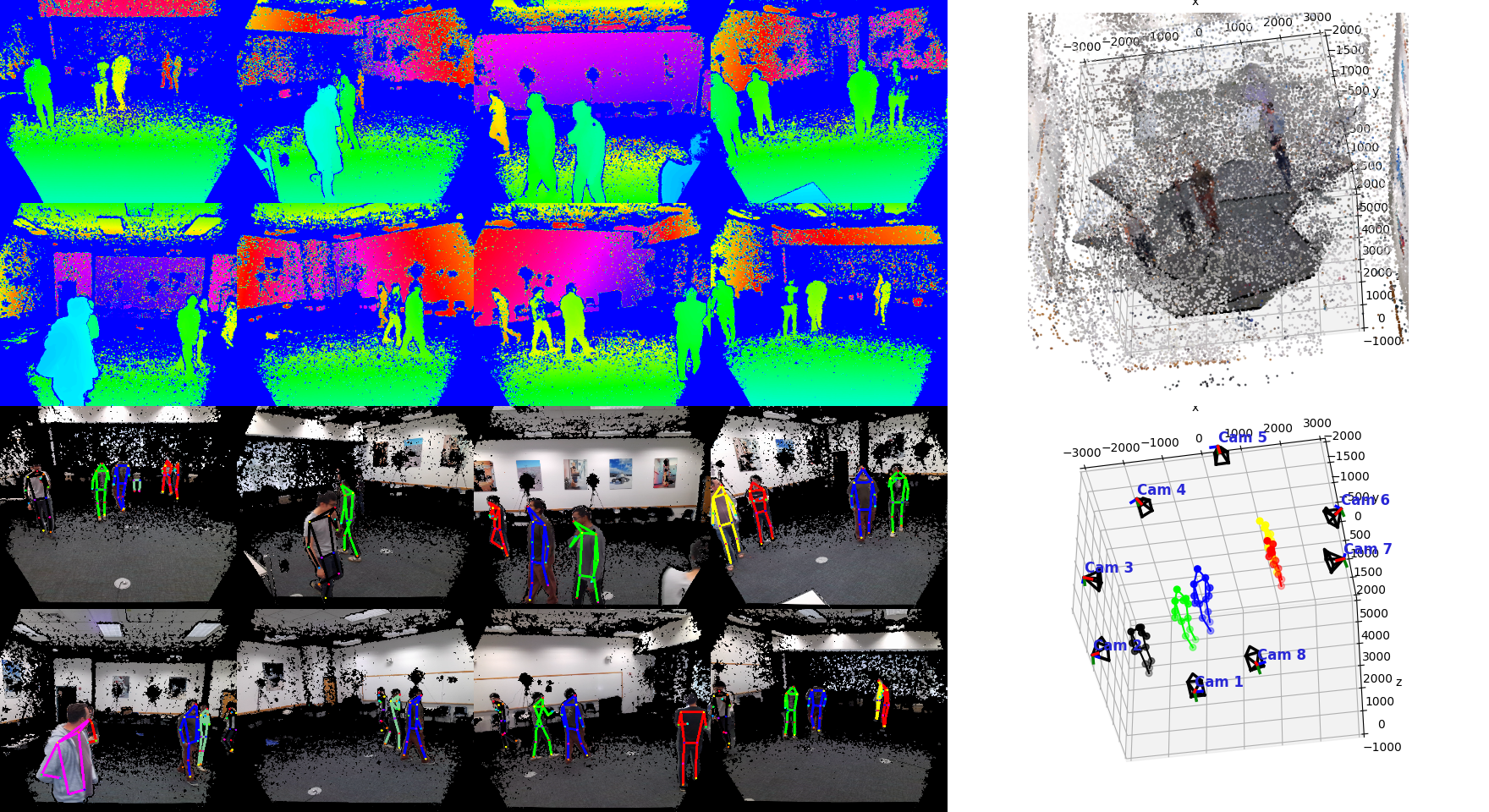}
  \vspace{-5mm}
  \caption{Qualitative example of 3D uncalibrated pose estimation with 8 cameras.}
  \label{fig:pose_8}
  \vspace{-4mm}
\end{figure*}

Besides the datasets in the main paper, we also have more challenging frames with people having interactions using our classroom V2 with \textbf{8} cameras. The dataset simulates two to three groups of people chatting with each other and have 7 sets each with 1000 frames (ending up with 7000 frames). We manually labeled 100 frames per set for evaluation and present the results in Table~\ref{tab:pcp_rebut}. We presented two examples of people with interactions in Figure~\ref{fig:pose_8}.

\begin{table*}[t!]
\small
\caption{\textbf{Comparisons on our collected datasets: Classroom v2.} The reported numbers are PCP values. The number in bold indicates the best results. }
\label{tab:pcp_rebut}
\centering
  \begin{tabular}{l|c|cccc|c}
    \toprule
    
    \multicolumn{1}{c}{Classroom v2} & \multicolumn{1}{|c}{Camera Pose} & \multicolumn{1}{|c}{Actor 1} & \multicolumn{1}{c}{Actor 2} & \multicolumn{1}{c}{Actor 3} & \multicolumn{1}{c}{Actor 4} & \multicolumn{1}{|c}{Average $\uparrow$}\\
    \midrule
     Belagiannis \etal~\cite{belagiannis20153d}& \checkmark & 81.4 & 80.6 & 75.7 & 83.2 & 80.2\\
     Ershadi \etal~\cite{ershadi2018multiple}& \checkmark & 83.5 & 82.1 & 77.7 & 84.5 & 82.0\\
     Dong \etal~\cite{dong2019fast}& \checkmark & 90.1 & 91.2 & 86.7 & 93.1 & 90.3\\
     UncaliPose~\cite{Xu_2022_BMVC} & - & 90.5 & 90.7 & 88.9 & 93.8 & 91.0\\
    \midrule
    MVD-HPE (Ours) & - & 92.2 & 91.5 & 89.8 & 94.2 & 91.9\\
    MVD-HPE (Ours) & \checkmark & \textbf{94.2} & \textbf{93.9} & \textbf{91.5} & \textbf{94.1} & \textbf{93.4}\\
    \bottomrule
  \end{tabular}
  \vspace{-6mm}
\end{table*}

\end{document}